\documentclass[letterpaper, 10 pt, journal, twoside]{IEEEtran}
\usepackage{amsmath,amsfonts}
\usepackage{algorithmic}
\usepackage{array}
\usepackage[caption=false,font=normalsize,labelfont=sf,textfont=sf]{subfig}
\usepackage{textcomp}
\usepackage{stfloats}
\usepackage{url}
\usepackage{verbatim}
\usepackage{graphicx}
\usepackage{cite}
\usepackage[ruled,vlined,linesnumbered]{algorithm2e}
\SetCommentSty{emph}
\usepackage{multirow}
\usepackage{graphicx} % for \graphicspath
\usepackage{url}
\usepackage{xcolor}
\usepackage{color}
\usepackage{amssymb,amsmath,comment}
\usepackage{amsfonts}
\usepackage[capitalize]{cleveref}
\usepackage{makecell}
\usepackage{booktabs}
\DeclareMathOperator*{\argmin}{arg\,min}
\Crefname{equation}{Eq.}{Eqs.}
\Crefname{figure}{Fig.}{Figs.}
\Crefname{tabular}{Tab.}{Tabs.}

\newtheorem{remark}{Remark}
\newtheorem{definition}{Definition}
\newcommand{\mymethod}{REMAP\xspace}

\newcommand{\mypredictor}{ExecTimeNet\xspace}
\newcommand{\myexecution}{ESADG\xspace}

\DeclareMathOperator*{\argmax}{arg\,max}

\begin{document}

% \title{From Discrete Plans to Real World Execution: \\ An Execution-Aware Framework for Multi-Agent Path Finding}
\title{From Discrete Plans to Real-World Execution:\\A World-Model-Driven Framework for Execution-Aware Multi-Agent Path Finding}

\author{
% {Anonymous Author(s)}
{Jingtian Yan,
Shuai Zhou,
He Jiang,
Stephen F. Smith,
Jiaoyang Li}
\thanks{This paper was produced by the IEEE Publication Technology Group. They are in Piscataway, NJ.}% <-this % stops a space
% \thanks{J. Yan, S. Zhou, S. Smith, and J. Li are with Robotics Institute, Carnegie Mellon University, Pittsburgh, PA 15213, USA {(email: jingtiay, shuaizho, ssmith@andrew.cmu.edu, jiaoyangli@cmu.edu)}.}
% \thanks{Digital Object Identifier (DOI): see top of this page.}
}

\markboth{Journal of \LaTeX\ Class Files,~Vol.~14, No.~8, August~2021}%
{Shell \MakeLowercase{\textit{et al.}}: A Sample Article Using IEEEtran.cls for IEEE Journals}


\maketitle

% Use action to represent the node in the ADG
% Use agent to represent  

\begin{abstract}
Multi-Agent Path Finding (MAPF) studies how to coordinate multiple agents to reach their goals without collisions and underpins a range of large-scale robotic systems, including automated warehousing and manufacturing. Recent advances enable MAPF solvers to compute high-quality plans for hundreds of agents. However, these plans are generated using simplified robot models with discretized time and action spaces. When they are deployed in physical systems, heterogeneous robot dynamics, asynchronous interactions, communication delays, and other real-world factors can lead to substantial deviations from planned performance.
We bridge the gap between discrete planning and real-world execution through \mypredictor, a learned world model of MAPF execution that predicts how a discrete MAPF solution will unfold on physical robots, mapping each discrete action to its realized execution state, including its wall-clock completion time and the kinodynamic state in which it ends.
Building on this capability, we first propose \mymethod, an execution-aware MAPF framework that integrates execution-time estimation into planning, guiding the search toward MAPF solutions with improved execution performance.
We also introduce \myexecution, a post-planning optimization procedure that optimizes the execution schedule of a given MAPF solution while preserving path feasibility.
We evaluate proposed frameworks in high-fidelity simulation with up to 300 agents and on physical robots.
In simulation, \mypredictor predicts the execution state accurately and transfers to unseen maps and agent counts beyond those seen in training.
Across simulation benchmarks spanning diverse map topologies, \mymethod reduces delays by up to 21\% over baselines, while \myexecution achieves up to 40\% normalized improvement.  
We also conduct a case study on MAPF with real-world deadlines, and combining both components lowers the deadline miss ratio by as much as 27\%.
On physical hardware, the full pipeline reduces total execution time by up to 15.3\%, demonstrating effective transfer from simulation to real-world deployment.
\end{abstract}


\section{Introduction}
\IEEEPARstart{M}{ulti}-Agent Path Finding (MAPF) seeks collision-free paths for a fleet of agents in a shared environment while optimizing objectives such as the sum of costs or makespan~\cite{Stern2019benchmark}. State-of-the-art MAPF solvers have demonstrated strong practical value across a range of robotics domains, including automated warehouses~\cite{wurman2008coordinating,honig2019warehouse}, traffic intersection coordination~\cite{li2023intersection,yan2024PSB}, and airport surface operations~\cite{morris2016planning}, where they can efficiently coordinate hundreds or even thousands of agents.

Despite this progress, a fundamental gap persists between the quality of plans produced by MAPF planners and their realized performance during execution. MAPF planners typically use simplified robot models and discrete timesteps. While these assumptions enable efficient planning, they do not capture the continuous and uncertain nature of real-world execution, where robots are subject to kinodynamic constraints, controller variability, and communication delays. As a result, MAPF plans cannot be executed directly without additional coordination mechanisms.
In practice, execution frameworks such as Action Dependency Graphs (ADG)~\cite{honig2019warehouse} and related representations~\cite{honig2016multi,berndt2023receding,yan2025winktpg} are used to transform discrete plans into executable schedules. These frameworks enforce collision avoidance and ensure safe and deadlock-free coordination under uncertainty by introducing precedence constraints between actions. While this layer enables robust execution, it also causes execution times to deviate from the estimated timesteps used during planning. A solution that is optimal under the planning model may therefore become suboptimal in practice, a phenomenon we refer to as the \emph{planning-execution gap}.

Several methods have been proposed to mitigate this gap. One direction integrates more accurate robot models directly into the planning process~\cite{honig2016multi,kottinger2022conflict,yan2025multi,moldagalieva2025db}.
Although such methods improve plan fidelity, they significantly increase computational complexity and are often limited in scalability.
Moreover, these models cannot fully capture real-world variability, as execution is influenced by additional factors such as controller noise, hardware differences, and communication delays.
% An alternative direction leverages data from real-world executions to predict execution-time behavior~\cite{ge2023congestion}. While promising, this prediction problem remains challenging. The completion time of an action depends not only on the local dynamics of the executing agent but also on delays accumulated along its preceding actions. In addition, interactions at shared locations couple agents through the dependency graph, allowing delays to propagate across multiple agents and over long dependency chains. These characteristics make accurate prediction difficult and limit the effectiveness of existing methods.

\begin{figure}
    \centering
\includegraphics[width=1.\linewidth]{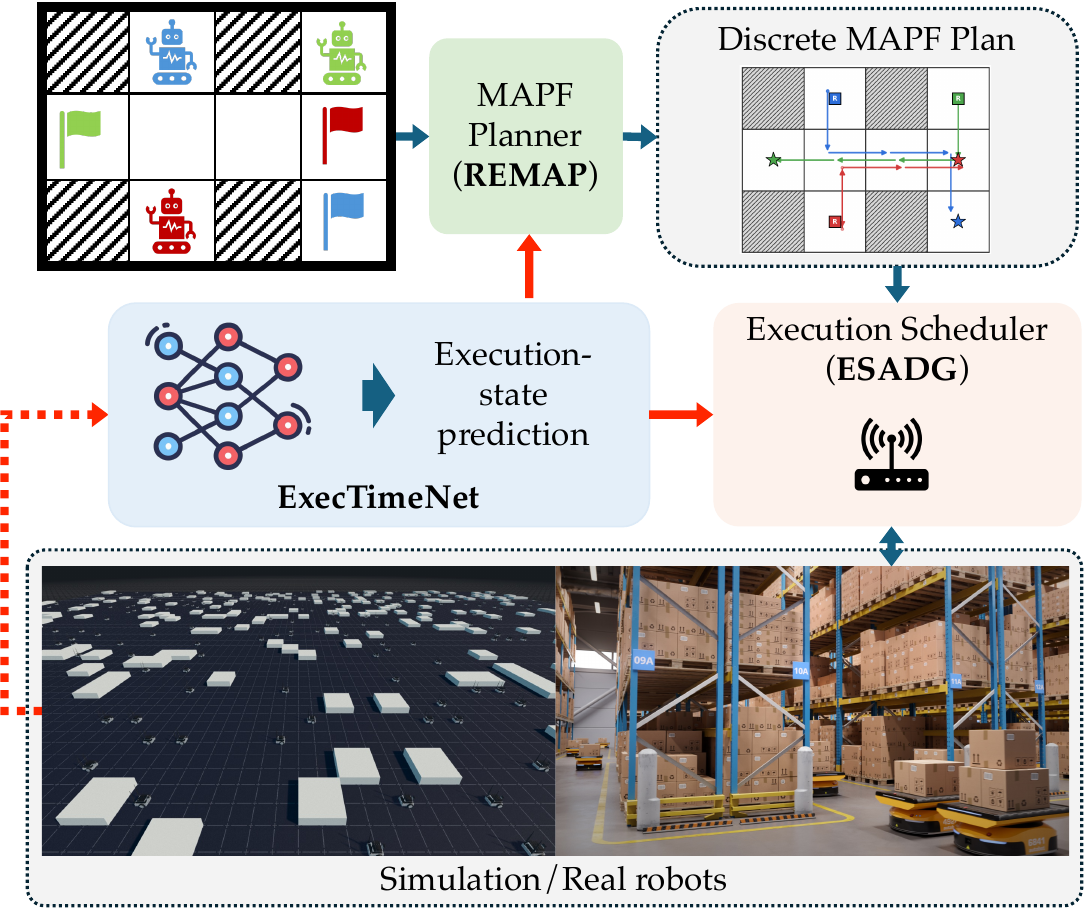}
\caption{Overview of our execution-aware MAPF framework. \textbf{Conventional pipeline (blue arrows).} A MAPF planner produces a discrete plan over the grid, which is handed to an execution layer that carries it out on real or simulated robots. The two stages are decoupled: the planner optimizes a discrete objective without knowing how the plan will actually unfold, so the realized execution can deviate substantially from the plan. \textbf{Our additions (red arrows).} We close this loop with \mypredictor, which predicts the execution state, including per-action completion times, from a candidate plan and feeds it back into both stages, making the planner (\mymethod) execution-aware during search and informing the execution scheduler (\myexecution) when ordering actions. The red dashed line denotes training: \mypredictor is learned offline from execution outcomes collected in the physics simulator and the real-robot warehouse testbed (bottom), which supply ground-truth execution state (e.g., completion times and speeds). Together, these paths bridge the gap between discrete planning and real-world execution.}
    \label{fig:intro_overview}
\end{figure}

To address the planning-execution gap, we propose a unified execution-aware MAPF framework that integrates execution information reasoning into both planning and execution as shown in~\cref{fig:intro_overview}.
At its core is \mypredictor, which acts as a \emph{world model} of MAPF execution: from a candidate plan it predicts the induced execution state, which both planning and execution-ordering optimization optimize against in place of costly simulation or physical rollouts~\cite{ha2018world,hafner2020dreamer}.
To our knowledge, this is the first framework to close the gap between planning and execution with a learned world model, embedding that model directly within the search loops of both planning and execution.
The main contributions of this paper are as follows.
(1) We propose \textbf{\mypredictor}, a learned world model of MAPF execution that predicts execution information for each agent from the dependency structure induced by a MAPF solution. The model captures both intra-agent temporal dependencies and inter-agent coordination effects and generalizes across environments and agent populations without retraining.
(2) We propose \textbf{\mymethod} (\textbf{R}ealistic \textbf{E}xecution-aware \textbf{M}ulti-\textbf{A}gent Path \textbf{P}lanning), a planning framework that integrates \mypredictor into the search process to guide planning toward solutions with improved execution performance. The framework requires only minimal modifications to existing search-based MAPF planners and is demonstrated with the popular planners CBS~\cite{sharon2015conflict} and MAPF-LNS~\cite{li2021anytime}.
% Since querying the predictor at every search step is accurate but slow, we introduce an adaptive gating mechanism that invokes \mypredictor only on candidates likely to benefit from execution-time estimation.
(3) We propose \textbf{\myexecution} (\textbf{E}xecution-aware \textbf{S}witchable \textbf{A}ction \textbf{D}ependency \textbf{G}raph), an anytime large neighborhood search (LNS) framework that optimizes the execution schedule of a MAPF solution.
Given a fixed MAPF plan, \myexecution operates in the execution stage, reordering the sequence in which agents traverse shared locations, with per-action predictions from \mypredictor steering the search toward feasible, deadlock-free schedules.
% Given a fixed MAPF plan, the order in which agents pass through shared locations has a substantial impact on execution time. \myexecution exploits this degree of freedom by iteratively reordering the passing sequence at shared locations, using per-action predictions from \mypredictor both to select which orderings to modify and to score the resulting candidates, while ensuring path feasibility and deadlock-free execution throughout the search.

We evaluate the proposed frameworks both in high-fidelity physics-based simulation~\cite{yan2025advancing} and on physical robots.
Across simulation benchmarks spanning diverse map topologies and agent counts, \mypredictor achieves a mean absolute percentage error of 2-6\% across in-distribution and out-of-distribution settings.
\mymethod attains up to 21\% improvement in execution-aware cost over baselines, and \myexecution reaches a normalized improvement ratio of up to 40\% relative to state-of-the-art execution-schedule optimizers.
In a case study on MAPF with real-world deadlines, combining both components lowers the deadline miss ratio by as much as 27\% compared to planners that rely solely on discrete cost proxies. On physical hardware, the full pipeline reduces total execution time by up to 15.3\%, suggesting that the framework transfers well from simulation to real-world deployment.

The remainder of this paper is organized as follows. \Cref{sec:background} introduces the background and reviews related work. \Cref{sec:execTimeNet} presents \mypredictor. \Cref{sec:remap,sec:ksadg} describe \mymethod and \myexecution, respectively. \Cref{sec:experiment} reports experimental results, and \cref{sec:conclusion} concludes.

\section{Background and Related Work} \label{sec:background}
This section first reviews Multi-Agent Path Finding (MAPF), then discusses how MAPF plans are executed in real-world deployments, and finally surveys existing efforts to bridge the planning-execution gap. Related work is discussed alongside each topic.

\subsection{Multi-Agent Path Finding} \label{sec:bg:mapf}
We define the Multi-Agent Path Finding (MAPF) problem on an undirected graph $G = (V, E)$ with vertex set $V$ and edge set $E \subseteq V \times V$.
We are given $M$ agents $A = \{a_1, \dots, a_M\}$, each with a start vertex $s_i \in V$ and a goal vertex $g_i \in V$.
Time is discretized into uniform steps.
A path for agent $a_i$ is a sequence of vertices
$\pi_i = (v_i^0, v_i^1, \dots, v_i^{T_i})$
with $v_i^0 = s_i$, $v_i^{T_i} = g_i$, and $(v_i^t, v_i^{t+1}) \in E$ or $v_i^t = v_i^{t+1}$ for all $t$.
A MAPF solution $\pi = \{\pi_1, \dots, \pi_M\}$ must be collision-free:
no two agents may occupy the same vertex at the same time step (vertex conflict) or traverse the same edge in opposite directions at the same time step (edge conflict).
Classical MAPF minimizes a discrete objective $J(\pi)$ such as the sum of path lengths $\sum_i T_i$ or the makespan $\max_i T_i$.

Two families of search-based MAPF solvers dominate the search for high-quality solutions.
The Conflict-Based Search (CBS) family~\cite{sharon2015conflict,li2021eecbs,LiIJCAI19} resolves collisions by incrementally adding constraints, producing optimal or bounded-suboptimal solutions. Its constraint-based structure offers strong theoretical guarantees and readily accommodates richer requirements such as task precedence constraints and time constraints.
The Large Neighborhood Search (LNS) family~\cite{li2021anytime,HuangAAAI22,li2022mapf} instead begins from a feasible solution and iteratively replans subsets of agents, achieving strong empirical performance and anytime behavior. It underlies competition-winning MAPF solvers and provides the refinement mechanism in state-of-the-art systems such as WPPL~\cite{Jiang2024Competition} and LaCAM3~\cite{jain2026graph}, where it substantially improves solution quality.
Beyond minimizing path cost, several works extend MAPF with explicit temporal requirements. MAPF with Deadlines (MAPF-DL)~\cite{ma2018mapfdl} has been addressed via network flow and integer linear programming~\cite{ma2018mapfdl,wang2022multirobot}, adapted search-based methods~\cite{ma2018mapfdl,LiICAPS21}, and extensions incorporating task assignments~\cite{huang2023deadline,fine2023anonymous}. 
Airport surface operations impose hard departure-time constraints on taxiing aircraft~\cite{morris2016planning,li2019departure}, and intersection coordination requires vehicles to clear shared space within tight timing windows~\cite{li2023intersection}.

\subsection{MAPF Plan Execution} \label{sec:bg:gap}
Standard MAPF planners typically rely on simplified robot models during planning, assuming infinite acceleration and perfect execution.
In practice, however, the agents are physical robots that execute these plans in continuous space and time.
Following common practice in automated warehouses and prior MAPF deployments~\cite{wurman2008coordinating,honig2019warehouse,berndt2023receding}, we target automated guided vehicles (AGVs): a differential-drive robot in a 2D workspace that translates forward, rotates in place, or waits, executing each motion through an onboard controller.
During execution, these motions are subject to kinodynamic constraints, such as velocity, acceleration, and turning limits, and to system uncertainties, including communication delays and controller variability.
As a result, the MAPF solution cannot be followed exactly during deployment.
% Nevertheless, agents must still satisfy the coordination constraints imposed by the MAPF solution, such as synchronizing actions at shared locations, to guarantee safe and deadlock-free execution.
Existing methods typically rely on either synchronized execution with online replanning or dedicated execution frameworks.

Synchronized execution enforces the discrete timing assumptions of the MAPF planner by requiring agents to progress in lockstep. While this guarantees collision-free execution, it forces the entire fleet to operate at the pace of the slowest agent, resulting in substantial idle time. 
Online replanning methods attempt to mitigate this limitation by recomputing plans reactively during execution.
For example, Time-independent planning~\cite{okumura2021time} replans one step at a time to absorb asynchronous delays, other methods repair coordination in response to disturbances~\cite{coskun2019repairMAPF} or stochastic travel times~\cite{kita2023online}.
While these methods improve responsiveness to execution deviations, most still rely on the same simplified motion assumptions as standard MAPF formulations. In addition, repeated replanning requires timely synchronization and communication among agents, making performance sensitive to communication delays.

To address these limitations, a second line of work employs dedicated execution frameworks~\cite{vcap2016provably,honig2019warehouse}. Given a MAPF solution, these frameworks construct a directed graph over agent actions to encode precedence constraints among agents. Such frameworks decouple high-level path planning from low-level execution coordination, enabling robust deployment under realistic system constraints. Although they differ in implementation details, these methods share a common underlying structure and can often be inter-translated.
This line of work is also the most widely adopted in real-world deployments, owing to its robustness to temporal delays and its support for asynchronous execution.
We adopt the Action Dependency Graph (ADG)~\cite{honig2019warehouse} as the canonical formulation throughout this paper.

\begin{figure}
    \centering
    \includegraphics[width=\linewidth]{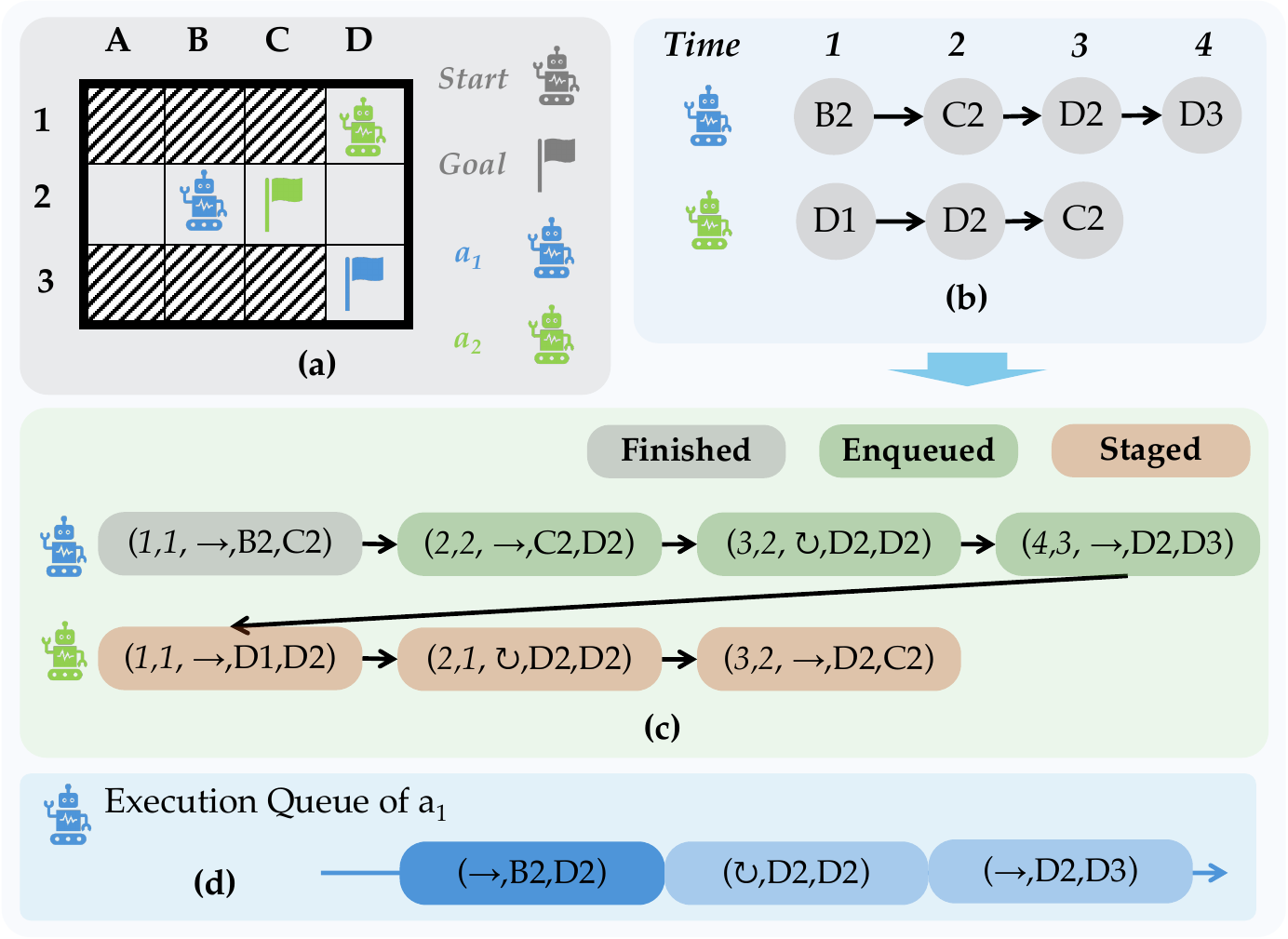}
    \caption{Illustration of the ADG. (a) A MAPF instance with two agents, $a_1$ and $a_2$. (b) The corresponding MAPF plan, giving each agent's location at every timestep. (c) The ADG built from the plan. Each node is an action written as $(k, t, \text{type}, \text{from}, \text{to})$, where $k$ is the action index, $t$ is the timestep in the original MAPF plan, \text{type} is the action ($\rightarrow$ translation, $\circlearrowright$ rotation), and the last two entries are its start and goal locations. Horizontal edges are Type-1, and vertical edges are Type-2. (d) Execution queue of $a_1$, where consecutive translation actions are merged into one longer action to avoid stop-and-go.}
    \label{fig:example_adg}
\end{figure}

\begin{definition}[Action Dependency Graph] \label{def:edg}
Given a MAPF solution $\pi$, an Action Dependency Graph (ADG) is a directed acyclic graph
$\mathcal{G}(\pi) = (\mathcal{V}, \mathcal{E}_1 \cup \mathcal{E}_2)$,
where each node $v_i^{m} \in \mathcal{V}$ denotes the $m$-th action of agent $a_i$, indexed by its position $m$ in the agent's action sequence; we separately record the discrete timestep at which the original MAPF plan scheduled the action. Each action is an atomic action (translation or rotation).
\emph{Type-1 edges} $\mathcal{E}_1$ connect consecutive actions of the same agent, preserving their sequential order.
\emph{Type-2 edges} $\mathcal{E}_2$ connect actions of different agents at shared locations, enforcing passing-order constraints: an edge $(u, v)$ indicates that the action $v$ may begin only after the action $u$ has completed execution.
During execution, each node carries one of three statuses, \emph{staged}, \emph{enqueued}, or \emph{finished}, initialized to \emph{staged}.
A node becomes \emph{enqueued} and is sent to the robot for execution once (1) it has no Type-1 predecessor or that predecessor is \emph{enqueued} or \emph{finished} and (2) all its Type-2 predecessors are \emph{finished}.
To enable higher operation speeds, consecutive enqueued actions of the same type are merged into one action during execution, avoiding stop-and-go between them.
A node is marked \emph{finished} when the robot confirms completion.
Execution ends when every node is \emph{finished}.
\end{definition}

\begin{table}[t]
\centering
\caption{Representative execution frameworks that encode MAPF solutions as dependency graphs.
Each framework represents either actions or locations as graph nodes and uses inter-agent edges to enforce passing orders at shared resources.}
\label{tab:dependency_frameworks}
\resizebox{\linewidth}{!}
{\begin{tabular}{l c c}
\toprule
Framework & Node Representation & Reorderable Inter-Agent Edges \\
\midrule
ADG~\cite{honig2019warehouse} & Actions & No \\
TPG~\cite{honig2016multi} & Locations & No \\
SADG~\cite{berndt2023receding} & Actions & Yes \\
STPG~\cite{feng2024real,JiangAAAI25} & Locations & Yes \\
BTPG~\cite{su2024bidirectional} & Locations & Yes \\
LDG~\cite{liu2024multi} & Locations & Yes \\
\bottomrule
\end{tabular}}
\end{table}

ADG always forces the original execution ordering induced by the MAPF plan, which can introduce unnecessary waiting. Several subsequent methods search for alternative valid orderings to improve execution efficiency while maintaining safe execution.
SADG~\cite{berndt2023receding} formulates execution-order optimization as Mixed Integer Linear Programming (MILP), which yields optimal solutions under the discrete MAPF model but scales poorly to large instances. GSES~\cite{feng2024real,JiangAAAI25} replaces MILP with a dedicated search algorithm. While improving scalability over SADG, they remain computationally expensive for large systems. There are also non-optimal methods for better scalability, such as LDG~\cite{liu2024multi}, which reduces waiting online, and BTPG~\cite{su2024bidirectional}, which enables first-come-first-served execution. As summarized in \Cref{tab:dependency_frameworks}, these execution frameworks use the same underlying structure.

% While these methods provide a robust framework to execute MAPF plans, such methods often suffer from a mismatch in performance between MAPF plan and execution.
% A plan that is optimal in discrete cost can therefore produce substantial degradation during execution.

While these frameworks enable robust execution, both the planning and execution-order optimization aim at an objective that diverges from realized execution time.
On the planning side, a recent study shows that minimizing a discrete objective in MAPF does not consistently translate into better execution performance, exposing a gap that an execution-aware objective is needed to close~\cite{yan2025analyzing}.
On the execution side, the order-optimization frameworks above minimize a completion time computed from a similarly simplified model rather than the true execution outcome, inheriting the same mismatch.
Both observations motivate our central idea of predicting realized execution time directly and using it to guide both planning and execution-order optimization.

\subsection{Execution-Time Modeling and Prediction}
One method of reducing the planning-execution gap is to improve the fidelity of the robot model used during planning. Some methods incorporate more accurate robot dynamics directly into MAPF planning~\cite{kottinger2022conflict,yan2025multi}, allowing solutions that better reflect physical execution constraints.
However, they typically incur substantially higher computational complexity while still failing to capture execution uncertainty arising from robot interactions, controller variability, and communication delays.

Recent work has explored learning-based methods to predict execution-time behavior from execution data. Examples include congestion prediction for large robot fleets~\cite{ge2023congestion} and adaptive planners that update travel-time estimates from execution feedback~\cite{kita2023online}. Similar ideas have also been studied in related domains, including highway traffic forecasting~\cite{song2020spatial}, train delay prediction~\cite{wen2023enough}, and ETA prediction in large-scale navigation systems~\cite{derrow2021eta}.
While promising, this prediction problem remains challenging.
% The completion time of an action depends not only on the local dynamics of the executing agent but also on delays accumulated along its preceding actions. In addition, interactions at shared locations couple agents through the dependency graph, allowing delays to propagate across multiple agents and over long dependency chains. These characteristics make accurate prediction difficult and limit the effectiveness of existing methods.
Existing methods primarily focus on local-region or per-agent features and do not explicitly model the dependency structure through which execution delays propagate across agents.
However, in MAPF, The completion execution state of an action depends not only on the executing agent's local dynamics but also on delays accumulated along its preceding actions, and interactions at shared locations couple agents so that a single disturbance can cascade across many agents and over long dependency chains.
In contrast, our method models both inter-agent dependencies, the propagation of delays across robots at shared locations, and intra-agent dynamics, the timing and kinodynamic state along each robot's path, through a learned model.

\section{Execution-Time Prediction (\mypredictor)}\label{sec:execTimeNet}

\begin{figure*}[t!]
    \centering
    \includegraphics[width=\linewidth]{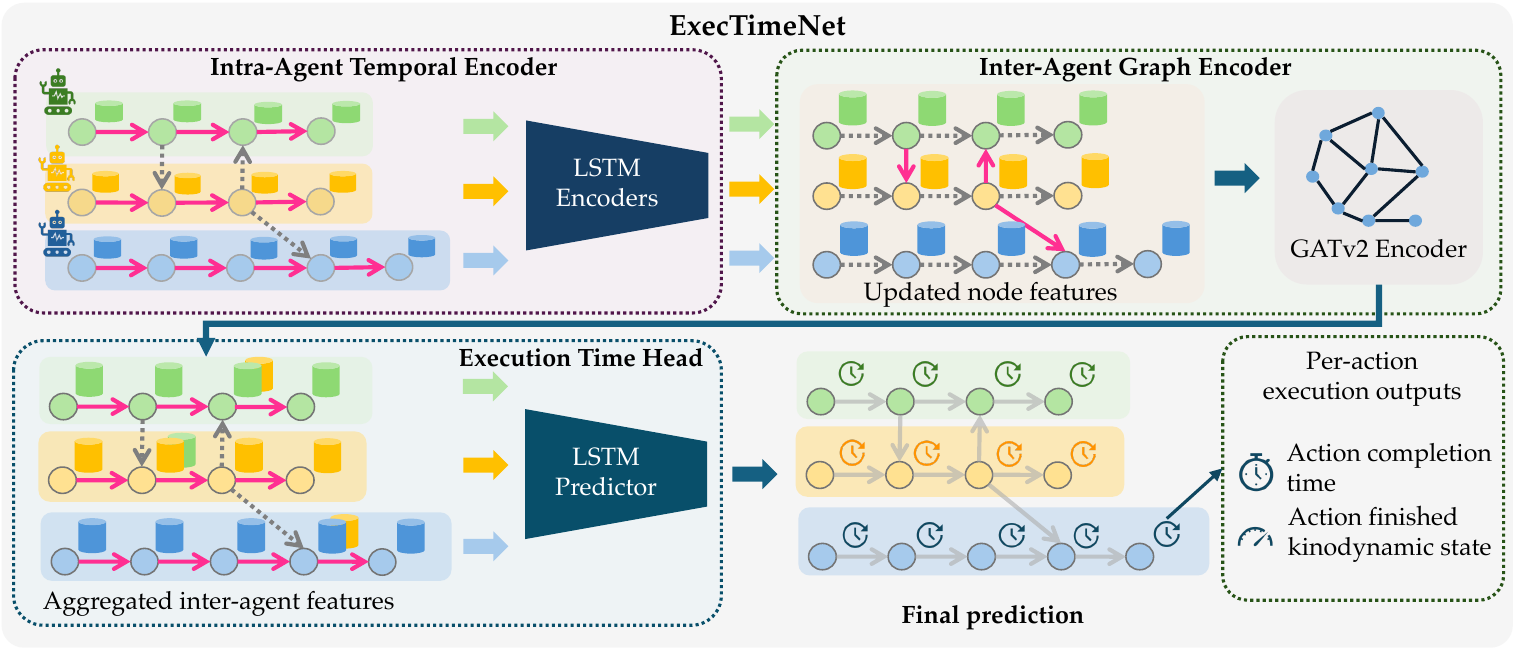}
    \caption{Model structure of \mypredictor. Pink arrows indicate the direction of information propagation. The intra-agent temporal encoder is a single weight-shared LSTM applied independently to each agent's sequence.}
    \label{fig:method_predictor}
\end{figure*}

In this section, we introduce \mypredictor, which predicts execution information from a discrete MAPF plan.
The remainder of this section is organized as follows. We first state
the problem formally (\Cref{sec:execTimeNet:problem}), then describe the
graph encoding used to represent an ADG as neural-network input
(\Cref{sec:execTimeNet:encoding}), and finally present the \mypredictor
architecture (\Cref{sec:execTimeNet:model}).

\subsection{Problem Statement}\label{sec:execTimeNet:problem}
Given a MAPF solution $\pi$, the goal of the \emph{execution-time prediction problem} is to predict the wall-clock completion time of every action in the plan.
During execution, each agent follows a continuous trajectory subject to its own kinodynamic constraints, e.g., individual bounds on velocity, acceleration, and angular velocity, and must wait before entering the shared locations according to the precedence constraints encoded in the ADG.
These factors cause the actual completion time of each action to deviate from its scheduled discrete timestep in ways that depend on the individual path geometry, the agent's kinodynamic parameters, and the global dependency structure $\mathcal{G}(\pi)$.

Formally, let $\mathcal{U} = \{u_1, \dots, u_M\}$ denote the per-agent kinodynamic parameters, where $u_i$ specifies the velocity, acceleration, and angular velocity limits of agent $a_i$.
Given an ADG $\mathcal{G}(\pi) = (\mathcal{V}, \mathcal{E}_1 \cup \mathcal{E}_2)$ constructed from $\pi$ together with $\mathcal{U}$, the prediction problem is to learn a function
\[
\Phi:\bigl(\mathcal{G}(\pi),\,\mathcal{U}\bigr) \;\longrightarrow\; \{(\hat{\tau}_v, \hat{s}_v)\}_{v \in \mathcal{V}}
\]
that estimates, for each action $v \in \mathcal{V}$, the completion time $\hat{\tau}_v$ and state $\hat{s}_v$ during execution.
Each agent's goal arrival time is recovered as the predicted completion time of its final action. More generally, $\Phi$ could be extended to output a probability distribution $p_v(t)$ over completion times to capture the stochastic variability inherent in physical systems.

\subsection{Graph Encoding}\label{sec:execTimeNet:encoding}

Given a MAPF solution, we construct its ADG and attach features to nodes, edges, and agents, producing an annotated graph that serves as input to \mypredictor. The features are grouped into four categories, each targeting a distinct source of execution-time variation. 
\emph{Action features} describe the local semantics of each action, since translation and rotation incur markedly different execution costs under kinodynamic constraints. Each node carries a binary indicator distinguishing the two action types, the rotation magnitude, the translation distance, and the discrete timestep assigned by the MAPF plan.
\emph{Graph structural features} encode the topological role of each node, which determines how delays propagate. We include the action's index along its agent's path, the in-degree and out-degree over Type-2 edges, the cumulative counts of preceding and subsequent translations of the same agent, the cumulative counts of preceding and subsequent rotations, and a binary indicator of whether the node has any outgoing inter-agent edges and therefore blocks actions of other agents.
\emph{Edge features} expose the temporal gap between connected actions, which is informative for predicting waiting durations.
Each edge carries its type, the timestep difference between the connected actions from the original MAPF solution, and the difference in their indices from ADG.
\emph{Per-agent kinodynamic features} support fleets with heterogeneous motion capabilities. Each agent is annotated with its maximum linear velocity, maximum linear acceleration, and maximum angular velocity, allowing \mypredictor to condition on individual limits rather than assuming a uniform platform.

\subsection{\mypredictor Model}\label{sec:execTimeNet:model}
Given the graph encoded from the ADG, predicting execution times presents two main challenges.
First, it introduces complex temporal dependencies along Type-1 edges:
actions at earlier nodes can propagate delays that affect the execution times of actions at later nodes.
Second, it introduces complex inter-agent dependencies along Type-2 edges: the execution of one agent's action directly constrains the timing of another's.
These dependencies are often non-local and span multiple hops in the graph.
Thus, traditional sequential or time-series prediction models~\cite{ge2023congestion} are inadequate.
To address these challenges, we propose \mypredictor, as shown in~\cref{fig:method_predictor}.
The model consists of three components: (1) an \emph{Intra-Agent Temporal Encoder} that captures the long-range temporal dependencies within each agent’s action sequence, (2) an \emph{Inter-Agent Graph Encoder} that captures dependencies across agents using the graph structure, and (3) an \emph{Execution Time Head} that generates per-action execution time and kinodynamic-state predictions.

% \subsubsection{Model Overview}  
Given the encoded graph, we first use the \emph{Intra-Agent Temporal Encoder} to capture temporal dependencies for the action sequence of each agent. Concretely, we treat the sequence of nodes corresponding to a single agent as an ordered trajectory and apply an LSTM~\cite{hochreiter1997long} encoder to update the features of each node by attending to all other actions from the same agent.
This enables the model to capture long-range temporal dependencies, regardless of their distance in the graph.
Next, we pass the graph with updated node features to the \emph{Inter-Agent Graph Encoder}, which leverages a GATv2~\cite{brody2022how} to propagate information across edges while incorporating edge features. This allows the model to account for inter-agent interactions, where the execution of one agent’s action can affect the timing of others through direct or indirect interactions.
Finally, we use an \emph{Execution Time Head} to generate per-action predictions.
An LSTM predictor aggregates the fused features along each agent's action sequence and produces two outputs at every node: the action's wall-clock completion time and the kinodynamic state at which the action terminates, e.g., instantaneous linear and angular velocity.
The completion-time output supplies the temporal signal needed by planning and execution optimization, while the kinodynamic-state output captures how the agent enters the following action and provides auxiliary signal that stabilizes training.
The model is trained by minimizing the Mean Absolute Percentage Error (MAPE) between predicted and ground-truth completion times and states,
$\mathcal{L}
  =
  \frac{1}{|\mathcal{V}|}
  \sum_{v \in \mathcal{V}}
  \left(
  \frac{\lVert \tau_{v} - \hat{\tau}_{v} \rVert}{\lVert \tau_{v} \rVert}
  + 
  \frac{\lVert s_{v} - \hat{s}_{v} \rVert}{\lVert s_{v} \rVert}
  \right)$,
where $\tau_v$ and $s_v$ denote the ground-truth completion time and state of action $v$, $\hat{\tau}_v$ and $\hat{s}_v$ are the corresponding predictions.

\begin{figure*}
    \centering
    \includegraphics[width=\linewidth]{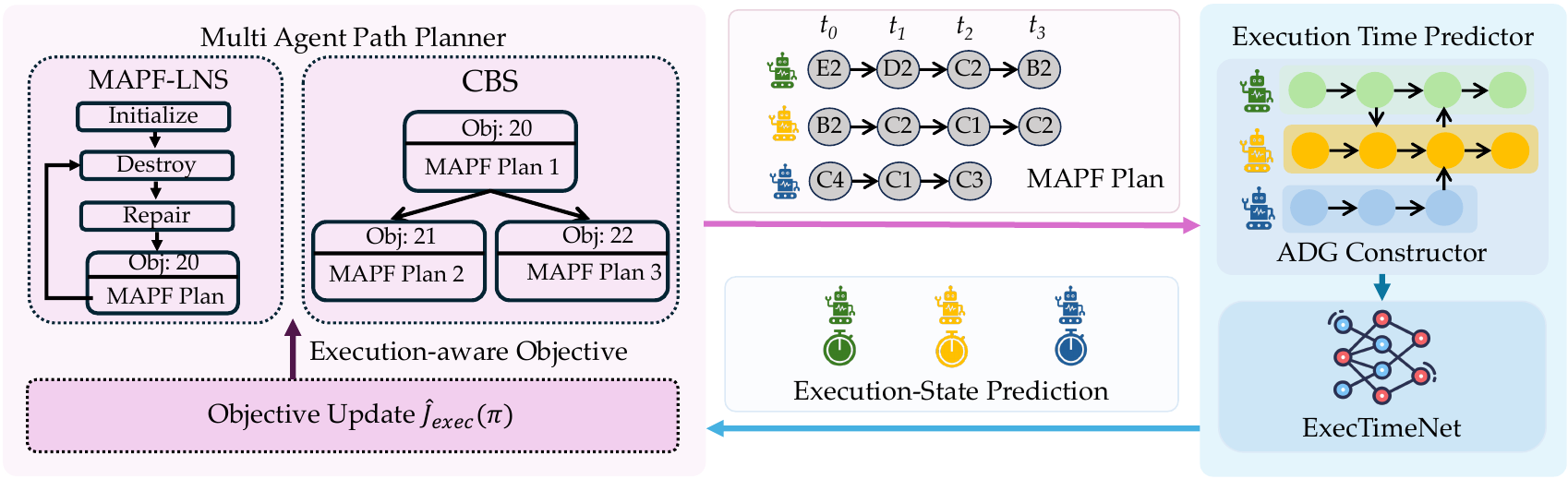}
    \caption{Overview of \mymethod.}
    \label{fig:frameworkstructure}
\end{figure*}

\section{Execution-Aware Planning (\mymethod)} \label{sec:remap}

This section introduces \mymethod, a framework that integrates \mypredictor into the search loop of existing MAPF planners to optimize execution-aware objectives.
We begin by formalizing execution-aware MAPF as a general optimization problem (\cref{sec:remap:problem}).
We then present the \mymethod framework and show that it integrates with both families of search-based solvers that achieve high-quality MAPF solutions, as identified in~\cref{sec:bg:mapf}.
We instantiate it on CBS~\cite{sharon2015conflict} and MAPF-LNS~\cite{li2021anytime} as representatives of these two families.
Finally, we introduce an online adaptive gating mechanism that reduces prediction overhead, illustrated through its integration with MAPF-LNS~\cite{li2021anytime} (\cref{sec:remap:lns}).

\subsection{Problem Statement}\label{sec:remap:problem}
The MAPF formulation in \cref{sec:bg:mapf} optimizes a
discrete objective $J_{\text{plan}}(\pi)$ defined over path
lengths $\{T_1,\dots,T_M\}$.
However, when agents execute their plans under realistic constraints, the actual execution time $\tau_i(\pi)\in\mathbb{R}_{+}$ at which agent $a_i$ reaches its goal can deviate substantially from $T_i$.
We extend MAPF by replacing the discrete planning objective with a function defined over these real execution times.
An \emph{execution-aware objective} is a function
\[
  J_{\text{exec}}(\pi)
  = F\!\big(\tau_1(\pi),\dots,\tau_M(\pi)\big)
\]
that maps the vector of per-agent execution times to a scalar cost.
Because $F$ is left unspecified, this formulation accommodates a range of practically relevant objectives, including minimizing execution makespan $\max_i\tau_i(\pi)$, reducing energy consumption, and enforcing deadline constraints.
For ease of exposition, we instantiate $F$ as the sum of execution times and use it as the running example throughout the method.

\begin{definition}[Execution-Aware MAPF]\label{def:ea-mapf}
Given a graph $G$, a set of agents $A$, and an execution-aware objective $J_{\text{exec}}$, find a collision-free plan
\[
  \pi^{*}=\argmin_{\pi\in\Pi}\;J_{\text{exec}}(\pi),
\]
where $\Pi$ is the set of feasible MAPF solutions.
\end{definition}

\subsection{Framework Overview} \label{sec:remap:cbs}
\mymethod integrates execution-time prediction into the search loop of existing MAPF planners with minimal modification.
The overall structure is shown in~\Cref{fig:frameworkstructure}.

Whenever the planner generates a candidate set of paths, \mymethod constructs the ADG, queries \mypredictor for per-agent execution time estimates $\hat{\tau}_i(\pi)$, and evaluates the predicted objective $\hat{J}_{\text{exec}}(\pi) = F(\hat{\tau}_1(\pi), \dots, \hat{\tau}_M(\pi))$, feeding the result back to guide the search.
Because the true execution times are unavailable during planning, $\hat{J}_{\text{exec}}$ serves as a surrogate for the objective $J_{\text{exec}}$ in~\Cref{def:ea-mapf}.
This interaction occurs at every search iteration, allowing the planner to compare candidate solutions using execution-aware costs rather than discrete path lengths.

\mymethod requires only that the underlying planner can produce candidate path sets during search, regardless of whether those solutions still contain collisions.
This capability is available in both CBS and LNS families and their variants, which together cover the majority of high-quality MAPF solvers.
% We demonstrate integrations with CBS~\cite{sharon2015conflict} and MAPF-LNS~\cite{li2021anytime} as representative examples.

\subsubsection{Integration with CBS}
CBS~\cite{sharon2015conflict} is a bi-level search that maintains a tree of constraint nodes.
Each node contains a set of constraints and a set of single-agent paths that satisfy them but may still contain collisions.
When a collision is detected, CBS branches into two child nodes, each adding a constraint to prevent one agent from occupying the conflicting location at the conflicting time.
Nodes are sorted by objective cost, and the search terminates when a collision-free node is expanded.
\mymethod integrates with CBS by querying \mypredictor at each node to replace the discrete path cost with predicted execution-time cost for node ordering.
Because an internal node's paths may still collide, we build its ADG directly over those paths, adding Type-2 edges only at collision-free shared locations, so \mypredictor can score the node without explicit conflict resolution.

\begin{figure}[t]
    \centering
    \includegraphics[width=.5\linewidth]{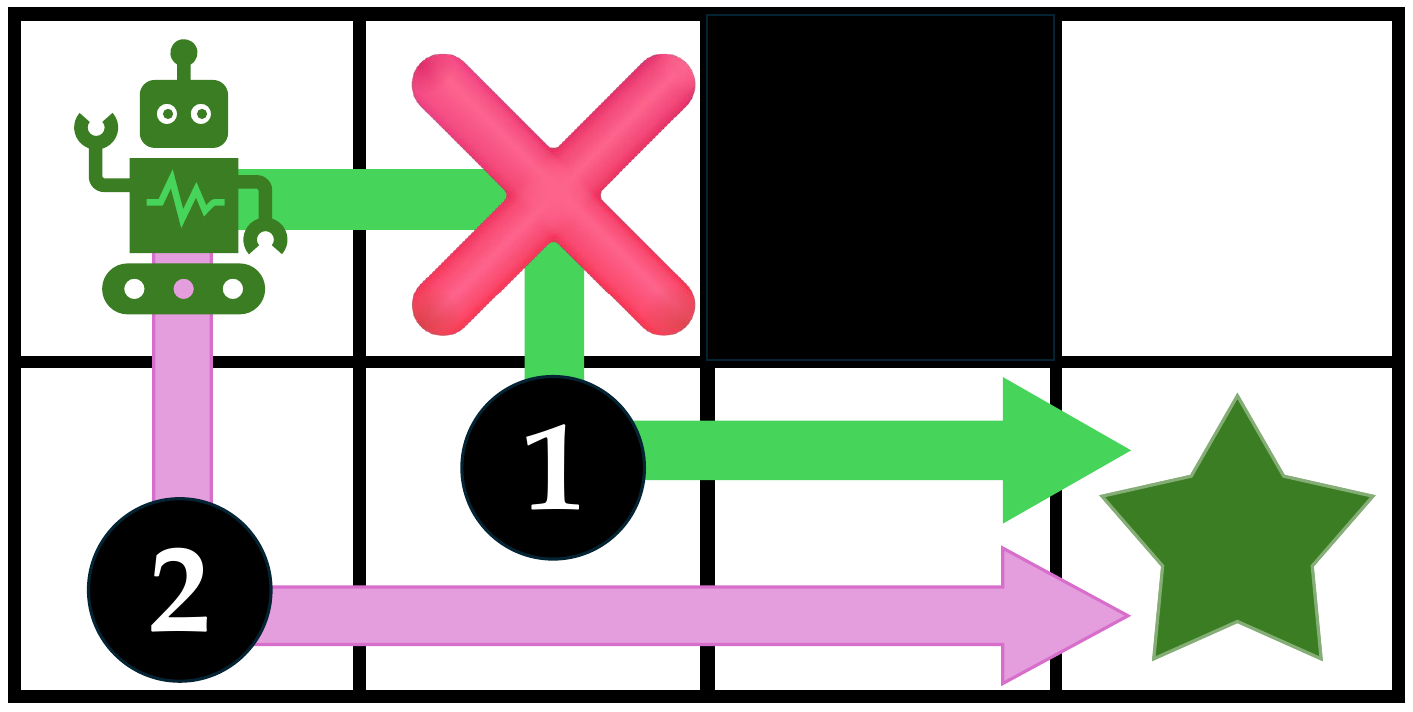}
    \caption{An example showing the loss of CBS optimality under execution-aware objectives. Paths~1 and~2 have equal discrete length, but path~2 requires fewer turns and executes faster.}
    \label{fig:CBS_toy}
\end{figure}

\begin{remark}[CBS Optimality] \label{remark:cbs_opt}
CBS guarantees optimality in standard MAPF because conflict resolution monotonically increases the sum of path costs~\cite{sharon2015conflict}.
This monotonicity breaks under execution-aware objectives: as shown in~\Cref{fig:CBS_toy}, two paths of equal discrete length can have different execution times (e.g., due to turning costs), so conflict resolution may \emph{decrease} $J_{\text{exec}}$.
Since the low-level planner optimizes path length rather than execution time, CBS is suboptimal for execution-aware MAPF.
\end{remark}

\subsubsection{Integration with MAPF-LNS}
MAPF-LNS~\cite{li2021anytime} is an anytime MAPF solver based on large neighborhood search.
Given an initial collision-free plan produced by a fast initial solver, it iteratively selects a neighborhood of agents, replans their paths while keeping the remaining agents fixed, and accepts the new paths only if they improve the planning objective.
Repeated over many iterations, this procedure produces a sequence of feasible solutions with monotonically non-increasing cost under the planning objective.
\mymethod integrates with MAPF-LNS by replacing the discrete sum-of-costs criterion used to accept neighborhood replans with the execution-aware objective $J_{\text{exec}}$.
After each replanning step, the updated plan is passed to \mypredictor to obtain per-agent execution-time estimates, which are aggregated into $J_{\text{exec}}$ and compared against the incumbent.

\subsection{Online Adaptive Gating} \label{sec:remap:lns}
The direct integration described above invokes \mypredictor at every accepted replan, which becomes expensive when neighborhoods are large or wall-clock budgets are tight.
The overhead is especially costly for iterative MAPF solvers such as MAPF-LNS, whose solution quality scales with the number of refinement iterations completed within a fixed time budget: every predictor call displaces a candidate replan that could otherwise improve the current solution.

A natural way to reduce this cost is to gate predictor calls using a cheap planner-side objective $J_{\text{fast}}$, such as the sum of costs, which can be computed directly from the candidate plan.
While $J_{\text{fast}}$ is not a faithful surrogate for $J_{\text{exec}}$ in general, prior work has shown that fast objectives of this form exhibit strong correlation with actual execution time~\cite{yan2025analyzing}.
This suggests that when the change $\Delta J_{\text{fast}}$ between the candidate and incumbent is large, a substantial improvement or degradation in discrete cost typically coincides with a shift of the same sign in execution cost.
We therefore only need to query \mypredictor for accurate predictions in the ambiguous region where $\Delta J_{\text{fast}}$ is small, since candidates of similar discrete cost can exhibit substantially different execution profiles owing to turning costs, cascading waits, and dependency-graph structure.
However, a fixed threshold on $\Delta J_{\text{fast}}$ cannot reliably delineate this boundary, since the appropriate cutoff varies across maps, agent densities, and search phases.
We therefore learn the relationship between $J_{\text{fast}}$ and $J_{\text{exec}}$ online from the predictor queries issued during search, yielding an adaptive gating mechanism that relies on $J_{\text{fast}}$ when its signal is strong and calls \mypredictor only in the ambiguous region.
The remainder of this subsection specifies the gating model and describes its instantiation within MAPF-LNS.

\subsubsection{Online Bayesian Gating Model}
To support gating decisions, we first establish a posterior predictive model that returns the probability that a candidate reduces $J_{\text{exec}}$, given its fast-objective change $\Delta J_{\text{fast}}$.
A simple Bayesian linear regression is well suited to this role: it captures the trend between the two objectives, quantifies uncertainty in the presence of limited data, and admits closed-form online updates.
To build this model, we maintain an oracle observation buffer $\mathcal{D}$ during the search.
Each time \mypredictor is queried, we record the observed pair
\begin{equation} \label{eq:oracle_obs}
\bigl(\Delta J_{\text{fast}},\; \Delta J_{\text{exec}}\bigr)
= \bigl(J_{\text{fast}}(\pi') - J_{\text{fast}}(\pi),\;
       J_{\text{exec}}(\pi') - J_{\text{exec}}(\pi)\bigr),
\end{equation}
where $\pi'$ is the candidate solution and $\pi$ is the current solution.
These pairs accumulate in a buffer $\mathcal{D} = \{(\Delta J_{\text{fast},i},\, \Delta J_{\text{exec},i})\}_{i=1}^{n}$ over the course of the search, and we model the relationship via Bayesian linear regression with unknown noise variance,
\begin{equation} \label{eq:blr}
\Delta J_{\text{exec}} = b_0 + b_1 \cdot \Delta J_{\text{fast}} + \epsilon, \qquad \epsilon \sim \mathcal{N}(0, \sigma^2),
\end{equation}
under a conjugate normal-inverse-gamma prior over $(b_0, b_1, \sigma^2)$.
The posterior is maintained via sufficient statistics, so each update requires only constant-time arithmetic and does not grow with the number of observations.

Given a candidate with observed $\Delta J_{\text{fast}}$, the posterior predictive model over $\Delta J_{\text{exec}}$ follows a Student-$t$ distribution, from which we compute the probability that execution cost improves,
\begin{equation} \label{eq:p_improve}
p_{\text{imp}} = P(\Delta J_{\text{exec}} < 0 \mid \Delta J_{\text{fast}},\, \mathcal{D}).
\end{equation}
This scalar quantity drives all subsequent gating decisions.
Because the model is fit online, it adapts to each instance without manual tuning.
On maps where the $J_{\text{fast}}$-to-$J_{\text{exec}}$ correlation is strong, the model converges to low residual variance and skips most oracle calls early in the search.
On congested maps where cascading waits weaken the correlation, the model retains higher uncertainty and continues to query the predictor more frequently, correctly reflecting that execution-time prediction is most valuable in precisely those settings.

\begin{algorithm}[t]
\caption{\mymethod-LNS with online adaptive gating}
\label{alg:adaptive_gating}
\KwIn{Initial solution $\pi^0$, parameters $\theta_{\text{acc}}$, $\theta_{\text{rej}}$, $k_{\text{init}}$, $N_{\text{ref}}$}
\KwOut{Optimized solution $\pi^*$}
 
$J_{\text{exec}} \gets$ \texttt{QueryOracle}($\pi^0$) \label{line:init_query}\\
$\pi^* \gets \pi^0$, $J^*_{\text{exec}} \gets J_{\text{exec}}$, $n_{\text{auto}} \gets 0, \mathcal{D} \gets \emptyset$ \\ % \tcp*{Oracle observation buffer}
\While{time budget not exhausted \label{line:loop_start}}{
    $\pi' \gets$ \texttt{GenerateCandidate}($\pi^*$) \label{line:generate}\\
    $\Delta J_{\text{fast}} \gets J_{\text{fast}}(\pi') - J_{\text{fast}}(\pi^*)$ \label{line:delta}\\
    \tcp{Decide whether to query the oracle}
    \uIf{$|\mathcal{D}| < k_{\emph{init}}$ \textbf{or} $n_{\emph{auto}} \geq N_{\text{ref}}$ \label{line:safeguard}}{
        $\text{decision} \gets$ \texttt{Oracle}
    }
    \Else{
        $p_{\text{imp}} \gets P(\Delta J_{\text{exec}} < 0 \mid \Delta J_{\text{fast}},\, \mathcal{D})$ \label{line:gate_start}\\
        \uIf{$p_{\emph{imp}} \geq \theta_{\emph{acc}}$}{
            $\text{decision} \gets$ \texttt{AutoAccept}
        }
        \uElseIf{$p_{\emph{imp}} \leq \theta_{\emph{rej}}$}{
            $\text{decision} \gets$ \texttt{AutoReject}
        }
        \Else{
            $\text{decision} \gets$ \texttt{Oracle} \label{line:gate_end}
        }
    }
    \tcp{Resolve decision and update state}
    \uIf{$\text{decision} =$ \texttt{Oracle} \label{line:update_start}}{
        $J'_{\text{exec}} \gets$ \texttt{QueryOracle}($\pi'$) \\
         $\Delta J_{\emph{exec}} \gets J'_{\text{exec}} - J^*_{\text{exec}}$ \\
        $\mathcal{D} \gets \mathcal{D} \cup \{(\Delta J_{\text{fast}},\, \Delta J_{\text{exec}})\}$ \\
        $n_{\text{auto}} \gets 0$ \\
        \If{$\Delta J_{\emph{exec}} < 0$}{
            $\pi^* \gets \pi'$, $J^*_{\text{exec}} \gets J'_{\text{exec}}$
        }
    }
    \uElseIf{$\text{decision} =$ \texttt{AutoAccept}}{
        $\pi^* \gets \pi'$, $n_{\text{auto}} \gets n_{\text{auto}} + 1$
    }
    \Else{
        $n_{\text{auto}} \gets n_{\text{auto}} + 1$ \label{line:update_end}
    }
}
\Return{$\pi^*$}
\end{algorithm}

\subsubsection{Integration with MAPF-LNS}
We instantiate the gating mechanism within MAPF-LNS~\cite{li2021anytime} as a representative anytime solver. The full procedure is summarized in~\Cref{alg:adaptive_gating}.
The algorithm takes an initial MAPF solution $\pi^0$ and four hyperparameters: two probability thresholds $\theta_{\text{acc}}, \theta_{\text{rej}}$ that govern when the gating model commits to an auto-decision, a warm-up size $k_{\text{init}}$ controlling how many oracle observations are required before gating activates, and a refresh interval $N_{\text{ref}}$ that bounds the number of consecutive auto-decisions.
We use the term \emph{oracle} to refer to \mypredictor in the context of the gating mechanism, reflecting its role as the high-fidelity but expensive source of execution-time estimates.
The initial solution $\pi^0$ is first evaluated by the oracle to obtain its execution cost $J_{\text{exec}}$ (Line~\ref{line:init_query}).
We then initialize the best-known solution $\pi^*$, its cost $J^*_{\text{exec}}$, observation buffer $\mathcal{D}$, and auto-decision counter $n_{\text{auto}}$ accordingly.
The main loop (Line~\ref{line:loop_start}) then refines $\pi$ until the time budget is exhausted.

Each iteration begins by generating a candidate solution $\pi'$, where MAPF-LNS selects a neighborhood of agents and replans their paths, after which $\Delta J_{\text{fast}}$ is computed against $\pi^*$ (Lines~\ref{line:generate}--\ref{line:delta}).
Next, we decide whether an oracle query is needed.
The oracle is consulted directly whenever either of two safeguards is triggered (Line~\ref{line:safeguard}): fewer than $k_{\text{init}}$ observations in $\mathcal{D}$, indicating that the model does not yet have enough data for the posterior predictive model, or more than $N_{\text{ref}}$ consecutive auto-decisions, preventing the cached baseline from drifting as $\pi^*$ evolves.
Otherwise, the gating model computes $p_{\text{imp}}$ and applies a decision rule with thresholds $\theta_{\text{rej}} < \theta_{\text{acc}}$ (Lines~\ref{line:gate_start}--\ref{line:gate_end}).
A candidate is auto-accepted when $p_{\text{imp}} \geq \theta_{\text{acc}}$ and auto-rejected when $p_{\text{imp}} \leq \theta_{\text{rej}}$; in the intermediate range, the candidate is sent to the oracle, since neither outcome is sufficiently certain to bypass execution-time prediction.

After the decision, the solver updates its state according to which branch was taken (Lines~\ref{line:update_start}--\ref{line:update_end}).
An oracle query produces $J'_{\text{exec}}$ for $\pi'$, after which the observation pair is appended to $\mathcal{D}$ and the auto-decision counter is reset.
The best-known solution $\pi^*$ is then updated if the new cost beats $J^*_{\text{exec}}$.
An auto-acceptance advances $\pi^*$ to $\pi'$ and increments the auto-decision counter, while an auto-rejection only increments the counter.
Each oracle query therefore serves a dual purpose, advancing both the search and the gating model, so the two improve together as iterations accumulate.
On timeout, the algorithm returns the best-known solution $\pi^*$.

\section{Execution Ordering Optimization (\myexecution)} \label{sec:ksadg}
Given a MAPF plan, execution frameworks such as ADG encode the passing order at shared locations through Type-2 edges, whose default orientation is inherited from the discrete plan (\Cref{sec:background}).
Because of the planning-execution gap, this default order is rarely optimal in wall-clock time, leaving the passing order at shared locations as a separate decision available at execution.
SADG~\cite{berndt2023receding} and STPG~\cite{feng2024real,JiangAAAI25} exploit this degree of freedom by reordering the passing sequence while preserving the deadlock- and collision-free guarantees of the underlying MAPF plan.
\myexecution searches over the same space of orderings but optimizes a learned execution-time objective from \mypredictor.
The remainder of this section formalizes the problem (\Cref{sec:ksadg:problem}), presents the search procedure (\Cref{sec:ksadg:tabu}), and details the edge group selection (\Cref{sec:ksadg:selection}) and cyclic ADG repair (\Cref{sec:ksadg:repair}).

\subsection{Preliminaries} \label{sec:ksadg:prelim}
We briefly recall the structures that \myexecution operates on, as in \Cref{sec:bg:gap}, and state the results from prior work that the algorithm relies on.

\subsubsection{Deadlock and acyclicity}
Execution of an ADG is deadlock-free if and only if the graph is acyclic~\cite{honig2019warehouse,berndt2023receding}.
A cycle in the precedence relation corresponds to a circular wait at execution time, in which every action in the cycle is blocked by another action in the same cycle.
This equivalence makes acyclicity the only feasibility condition that needs to be checked during ordering optimization, and motivates the cycle-detection-based repair procedure used in \Cref{sec:ksadg:repair}.

\subsubsection{Switchable edges}
Type-2 edges encode the passing order at shared locations. Except for those that terminate at goal vertices or originate from start vertices, they may be switched to encode the reversed order without affecting collision-free execution as long as acyclicity is preserved~\cite{berndt2023receding}.
We denote this set of switchable edges as $\mathcal{E}_{\text{sw}} \subseteq \mathcal{E}_2$, and write $\sigma$ for an assignment of orientations over $\mathcal{E}_{\text{sw}}$.
The induced ADG is $\mathcal{G}(\pi, \sigma)$, and the feasible joint orientation space
\begin{equation}
\Sigma(\pi) = \{\, \sigma \mid \mathcal{G}(\pi, \sigma) \text{ is acyclic} \,\}
\end{equation}
contains all assignments that preserve deadlock-free execution.
Each $\sigma \in \Sigma(\pi)$ defines a feasible execution schedule over the same set of spatial paths of the agent.

\subsubsection{Edge groups}
Not all switchable edges can be reversed independently.
Previous work \cite{berndt2023receding} observed that, in certain configurations of agent paths, a group of switchable edges must be reversed jointly to preserve acyclicity, and reversing a strict subset of them introduces a cycle.
Jiang et al.~\cite{JiangAAAI25} formalized this observation by defining two switchable edges as \emph{groupable} if, in every acyclic orientation of the dependency graph, the two edges are either both reversed or both fixed.
Groupability is an equivalence relation and partitions the switchable edges into disjoint \emph{maximal edge groups}, each of which must be reversed as a unit.
We treat each group as the basic switch unit in \myexecution.

\subsection{Problem Statement}\label{sec:ksadg:problem}
We define this problem as an extension of the Switchable Action Dependency Graph (SADG) formulation~\cite{berndt2023receding}, where we extend the orientation optimization problem to an execution-aware objective.
Each $\sigma \in \Sigma(\pi)$ induces an ADG $\mathcal{G}(\pi, \sigma)$, and the execution ordering optimization problem is
\begin{equation}
\sigma^{opt} = \argmin_{\sigma \in \Sigma(\pi)} J_{\mathrm{exec}}\big(\mathcal{G}(\pi, \sigma)\big),
\end{equation}
where $J_{\mathrm{exec}}$ evaluates the execution performance of $\mathcal{G}(\pi, \sigma)$.
This formulation is a path-restricted instance of execution-aware MAPF (\Cref{def:ea-mapf}): the paths are fixed, and only the dependency structure varies.

\subsection{Overview of \myexecution} \label{sec:ksadg:tabu}

\begin{algorithm}[t]
\caption{Overview of \myexecution}
\label{alg:tabu_search}
\KwIn{Initial configuration $\sigma^{0}$, switchable edge groups $\mathcal{E}_{\text{sw}}$, hyper-parameter $K$}
\KwOut{Optimized configuration $\sigma^{*}$}
$\sigma^{*} \gets \sigma^{0}$ \label{line:esadg_init_state}\\
$J^{*}_{\text{exec}} \gets$ \texttt{QueryPredictor}($\sigma^{0}$) \label{line:esadg_init_cost}\\
$\mathcal{T} \gets \emptyset$ \tcp{Tabu list of forbidden groups}
\While{time budget not exhausted \label{line:esadg_loop_start}}{
    $\mathcal{C} \gets \{\, g \mid g \in \mathcal{E}_{\text{sw}},\; g \notin \mathcal{T} \,\}$ \label{line:esadg_candidates}\\
    \If{$\mathcal{C} = \emptyset$}{
        \textbf{break}
    }
    $g_{\text{select}} \gets$ \texttt{SelectGroup}($\mathcal{C}$, $\sigma^{*}$, $K$) \label{line:esadg_select}\\
    \tcp{Reverse and create new configuration}
    $\sigma_{\text{new}} \gets$ \texttt{ReverseGroup}($\sigma^{*}$, $g_{\text{select}}$) \label{line:esadg_reverse}\\
    \If{\textbf{not} \texttt{RepairADG}($\sigma_{\text{new}}$) \label{line:esadg_repair}}{
        $\mathcal{T} \gets \mathcal{T} \cup \{g_{\text{select}}\}$ \tcp{Infeasible: forbid revisit}
        \textbf{continue}
    }
    $J'_{\text{exec}} \gets$ \texttt{QueryPredictor}($\sigma_{\text{new}}$) \label{line:esadg_query}\\
    \If{$J'_{\text{exec}} < J^{*}_{\text{exec}}$}{
        $\sigma^{*} \gets \sigma_{\text{new}}$, $J^{*}_{\text{exec}} \gets J'_{\text{exec}}$, $\mathcal{T} \gets \emptyset$ \label{line:esadg_best_update}
    }
    \Else{
        $\mathcal{T} \gets \mathcal{T} \cup \{g_{\text{select}}\}$ \label{line:esadg_reject}
    }
}
\Return{$\sigma^{*}$}
\end{algorithm}

The optimization in \cref{sec:ksadg:problem} is challenging in several aspects.
$\Sigma(\pi)$ grows combinatorially with the number of switchable edges, making exhaustive enumeration infeasible.
Moreover, given the computational overhead of an \mypredictor call, only a small number of candidates can be evaluated within the time budget.
An arbitrary reversal can introduce a cycle and violate the deadlock-free requirement, so feasibility must be verified before each evaluation.
\myexecution approaches the problem as an anytime local search that concentrates predictor calls on the orderings most likely to improve execution time, and returns the best ordering found within a fixed time budget.

The overall pipeline of \myexecution is shown in~\cref{alg:tabu_search}.
It takes an initial configuration $\sigma^{0}$, the set of switchable edge groups $\mathcal{E}_{\text{sw}}$, and a hyper-parameter $K$ as input, where $\sigma^{0}$ is constructed from the initial collision-free MAPF solution.
The initial configuration is evaluated by \mypredictor and serves as the best-known state $\sigma^{*}$ (Lines~\ref{line:esadg_init_state}--\ref{line:esadg_init_cost}), and the tabu list $\mathcal{T}$ is initialized empty.
The main loop (Line~\ref{line:esadg_loop_start}) then refines $\sigma^{*}$ until the time budget is exhausted.

Each iteration begins by gathering the set $\mathcal{C}$ of non-tabu switchable groups, exiting if no candidates remain (Line~\ref{line:esadg_candidates}).
Querying \mypredictor for every group is too expensive at scale. Instead, \texttt{SelectGroup} scores the candidates in $\mathcal{C}$ with the cheap gain estimator discussed in~\Cref{sec:ksadg:selection} and returns only the top-ranked edge group $g_{\text{select}}$ for full evaluation (Line~\ref{line:esadg_select}).
The selected group is reversed to produce a candidate configuration $\sigma_{\text{new}}$, which is then passed through \texttt{RepairADG} (\Cref{sec:ksadg:repair}) to verify acyclicity; on failure, the group is added to $\mathcal{T}$ and the iteration ends (Lines~\ref{line:esadg_reverse}--\ref{line:esadg_repair}).
Otherwise, $\sigma_{\text{new}}$ is scored by \mypredictor (Line~\ref{line:esadg_query}).
If the new cost $J'_{\text{exec}}$ improves on the best-known cost $J^{*}_{\text{exec}}$, the best-known state is updated and the tabu list is cleared, since the search landscape has shifted and previously unpromising groups may now be worth revisiting (Line~\ref{line:esadg_best_update}).
Otherwise, the group is added to $\mathcal{T}$ to prevent immediate re-selection (Line~\ref{line:esadg_reject}).
On timeout, the algorithm returns the current best configuration $\sigma^{*}$.

\subsection{Edge Group Selection} \label{sec:ksadg:selection}
Given a non-tabu candidate set $\mathcal{C} \subseteq \mathcal{E}_{\text{sw}}$, \texttt{SelectGroup} returns a single edge group whose reversal is expected to yield the largest improvement in $J_{\mathrm{exec}}$.
The candidate space grows quickly with the agent count, but only a small fraction of reversals improves execution time.
The reason is structural: reversing a Type-2 edge $u \to v$ changes the schedule only when $u$ is the predecessor that currently keeps $v$ waiting.
Otherwise the edge is slack, and reversing it will not improve $J_{\mathrm{exec}}$.
\texttt{SelectGroup} exploits this property in three steps.
First, each candidate $g \in \mathcal{C}$ is scored by a lightweight gain estimator $\Delta(g)$ computed from the current arrival times and local precedence structure, requiring no predictor call (\Cref{sec:ksadg:arr_update}, \Cref{sec:ksadg:edge_gain}).
The top-$K$ groups by $\Delta(g)$ are then retained, concentrating the search on the candidates most likely to reduce $J_{\mathrm{exec}}$.
Finally, one group is sampled uniformly from the top-$K$ and returned, which preserves the bias toward high-gain candidates while preventing the search from cycling on the same deterministic choice.

\subsubsection{Arrival-Time Update} \label{sec:ksadg:arr_update}
The gain estimator builds on the approximation of how arrival times change when a Type-2 edge is switched.
Each node $v$ in the current ADG has an arrival time $T_{\mathrm{arr}}(v)$ from the most recent \mypredictor query and a predecessor set $\mathrm{Pred}(v)$.
Its action duration is $T_{\mathrm{travel}}(v) = T_{\mathrm{arr}}(v) - \max_{q \in \mathrm{Pred}(v)} T_{\mathrm{arr}}(q)$, the time between the latest predecessor finishing and $v$ completing.
When a predecessor $u$'s finish time changes to $T_{\mathrm{arr}}'(u)$, the updated finish time of $v$ is
\begin{equation}\label{eq:arr_update}
    T_{\mathrm{arr}}'(v) = \max\!\left(\max_{q \in \mathrm{Pred}(v) \setminus \{u\}} T_{\mathrm{arr}}(q),\; T_{\mathrm{arr}}'(u)\right) + T_{\mathrm{travel}}(v),
\end{equation}
which adds $T_{\mathrm{travel}}(v)$ to whichever predecessor now constrains $v$ the most.
The recurrence holds $T_{\mathrm{travel}}(v)$ fixed and re-evaluates only the waiting component.
Applied iteratively, it propagates a single change along downstream nodes without invoking \mypredictor.

\subsubsection{Gain Estimate} \label{sec:ksadg:edge_gain}
Each edge group $g$ involves two agents $a_i$ and $a_j$, with every edge in $g$ pointing in the same direction (\Cref{sec:ksadg:prelim}).
Reversing $g$ therefore removes all of $g$'s constraints on $a_j$ at once and replaces them with the reversed constraints on $a_i$.
We capture this effect with two quantities: the time $\Delta^+(g)$ that $a_j$ saves at the shared locations, and the delay $\Delta^-(g)$ that $a_i$ incurs from the new reversed constraints.

\begin{figure}[t]
    \centering
    \includegraphics[width=.74\linewidth]{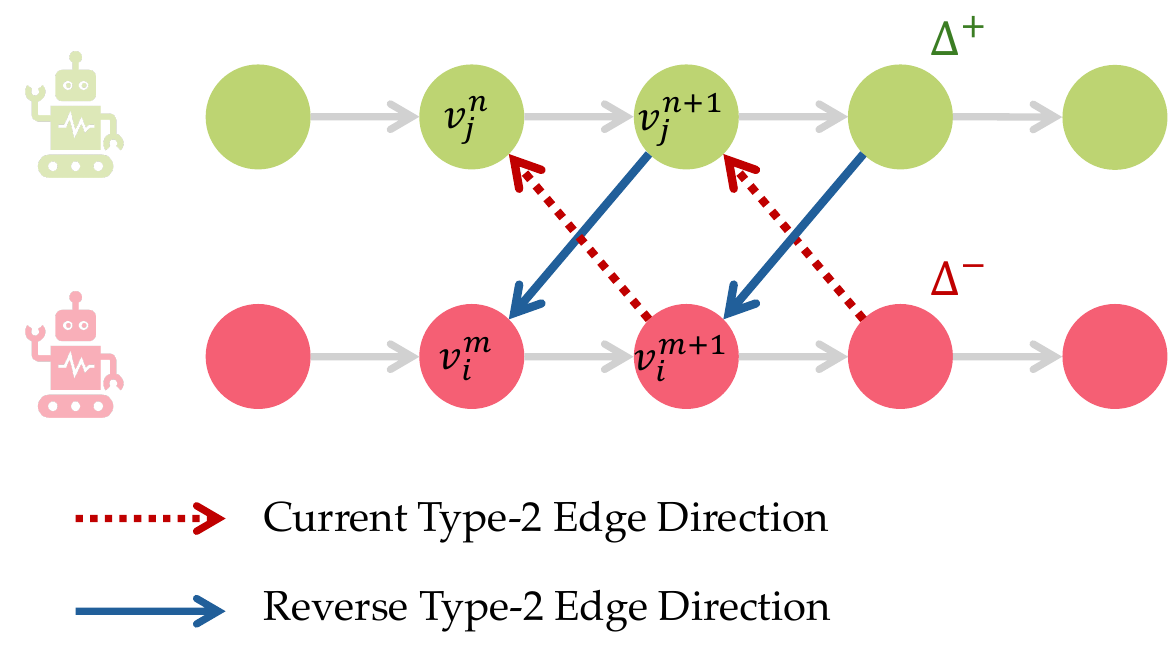}
    \caption{
    Illustration of the edge group selection gain estimator.
    Under the current Type-2 edge orientation (top), the red agent ($a_j$) waits at the shared location for the green agent ($a_i$) to pass.
    Reversing the edge (bottom) removes this wait, yielding a time reduction $\Delta^+$ for the red agent, while potentially introducing a delay $\Delta^-$ to the green agent.}
    \label{fig:heuristic_edge_selection}
\end{figure}

As shown in~\cref{fig:heuristic_edge_selection}, let $e^{*} = (v_i^{m+1}, v_j^{n})$ denote the edge in $g$ whose target has the smallest path index on $a_j$, i.e., the first location at which $a_j$ is blocked under the current orientation.
If $v_i^{m+1}$ is not the bottleneck predecessor of $v_j^{n}$, i.e., $T_{\mathrm{arr}}(v_i^{m+1}) < \max_{q \in \mathrm{Pred}(v_j^{n})} T_{\mathrm{arr}}(q)$, removing the group changes nothing along $a_j$'s path, and $\Delta^+(g) = 0$.
Otherwise, recomputing $T'_{\mathrm{arr}}(v_j^{n})$ via Eq. (6) with the predecessor $v_i^{m+1}$ removed and propagating the change along $a_j$'s path yields the saved time:
\begin{equation}
\Delta^+(g) = T_{\mathrm{arr}}(v_j^{T_j}) - T_{\mathrm{arr}}'(v_j^{T_j}),
\end{equation}
where $v_j^{T_j}$ is $a_j$'s final action.
The reversal symmetrically introduces $v_j^{n+1}$ as a new predecessor of $v_i^{m}$, so $a_j$ must clear the first shared location before $a_i$ enters.
Applying~\Cref{eq:arr_update} with $T_{\mathrm{arr}}'(v_j^{n+1})$ and propagating along $a_i$'s path yields
\begin{equation}
\Delta^-(g) = T_{\mathrm{arr}}(v_i^{T_i}) - T_{\mathrm{arr}}'(v_i^{T_i}) \;\leq\; 0,
\end{equation}
which is non-positive because the reversal can only add constraints to $a_i$.
The group score is
\begin{equation}
\Delta(g) = \Delta^+(g) + \Delta^-(g),
\end{equation}
positive when the time saved by $a_j$ outweighs the delay imposed on $a_i$.
\texttt{SelectGroup} ranks the non-tabu groups by $\Delta(g)$ in descending order, retains the top-$K$, and returns one sampled uniformly at random.

\subsection{Cyclic ADG Repair} \label{sec:ksadg:repair}

\begin{algorithm}[t]
\caption{\texttt{RepairADG}}
\label{alg:repair_adg}
\KwIn{Candidate configuration $\sigma$, switchable edge groups $\mathcal{E}_{\text{sw}}$, edge group gain estimator $\Delta(\cdot)$, max repair depth $D_{\max}$}
\KwOut{Repaired configuration $\sigma$ or \texttt{fail}}
$\mathcal{E}_{\text{visited}} \gets \emptyset$ \tcp{Visited groups}
\For{$d = 1$ \KwTo $D_{\max}$ \label{line:repair_loop_start}}{
    $\mathcal{E}_{\text{cycle}} \gets$ \texttt{DetectCycles}($\sigma$) \label{line:repair_detect}\\
    \If{$\mathcal{E}_{\text{cycle}} = \emptyset$}{
        \Return $\sigma$ \label{line:repair_success}
    }
    $\mathcal{C} \gets \{\, g \in \mathcal{E}_{\text{sw}} \mid g \in \mathcal{E}_{\text{cycle}},\; g \notin \mathcal{E}_{\text{visited}} \,\}$ \label{line:repair_candidates}\\
    \If{$\mathcal{C} = \emptyset$}{
        \Return \texttt{fail} \label{line:repair_fail_empty}
    }
    $g^{\star} \gets \argmax_{g \in \mathcal{C}} \Delta(g)$ \label{line:repair_select}\\
    $\sigma \gets$ \texttt{ReverseGroup}($\sigma$, $g^{\star}$) \label{line:repair_reverse}\\
    $\mathcal{E}_{\text{visited}} \gets \mathcal{E}_{\text{visited}} \cup \{g^{\star}\}$ \label{line:repair_visit}
}
\Return \texttt{fail} \label{line:repair_fail_depth}
\end{algorithm}

Reversing an edge group can introduce cycles, violating the acyclicity requirement for deadlock-free execution.
\Cref{alg:repair_adg} presents a lightweight repair procedure that iteratively resolves cycles by reversing additional edge groups, returning either a repaired configuration or \texttt{fail}.

The procedure takes the candidate configuration $\sigma$, the set of switchable edge groups $\mathcal{E}_{\text{sw}}$, the gain estimator $\Delta(\cdot)$ from \Cref{sec:ksadg:edge_gain}, and a hyper-parameter maximum depth $D_{\max}$.
A set $\mathcal{E}_{\text{visited}}$ tracks groups already reversed during repair, preventing the procedure from undoing its own progress.
Each iteration of the main loop (Line~\ref{line:repair_loop_start}) begins by detecting cycles in $\sigma$ via depth-first search and collecting their edges into $\mathcal{E}_{\text{cycle}}$ (Line~\ref{line:repair_detect}).
If $\mathcal{E}_{\text{cycle}}$ is empty, $\sigma$ is acyclic and repair succeeds (Line~\ref{line:repair_success}).
Otherwise, the candidate set $\mathcal{C}$ collects all unvisited switchable groups that share at least one edge with the detected cycles (Line~\ref{line:repair_candidates}); if no such group exists, no further reversal can break the remaining cycles and repair fails (Line~\ref{line:repair_fail_empty}).
The group with the highest gain estimate $g^{\star} = \argmax_{g \in \mathcal{C}} \Delta(g)$ is selected (Line~\ref{line:repair_select}), reversed in $\sigma$ (Line~\ref{line:repair_reverse}), and added to $\mathcal{E}_{\text{visited}}$ (Line~\ref{line:repair_visit}).
The procedure terminates with \texttt{fail} if the depth limit $D_{\max}$ is reached without restoring acyclicity (Line~\ref{line:repair_fail_depth}).

\section{Empirical Evaluation} \label{sec:experiment}

\begin{figure}
    \centering
    \includegraphics[width=1.\linewidth]{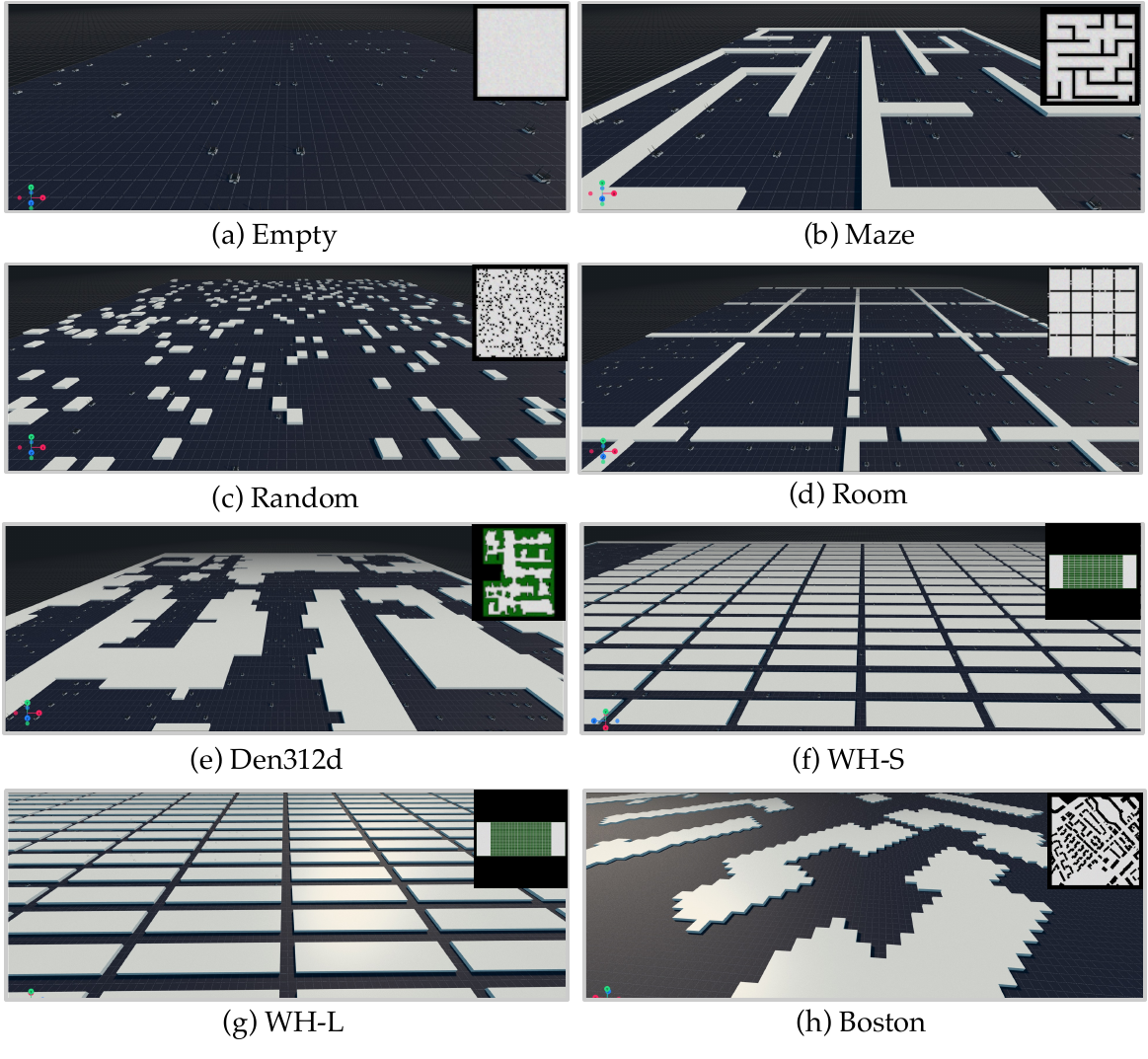}
    \caption{Eight physics-based simulation environments in SMART, derived from standard benchmark maps.}
    \label{fig:exp:sim_env}
\end{figure}

All experiments are conducted on an Ubuntu 22.04 system equipped with an Intel 13700K processor, 64\,GB of RAM, and an NVIDIA 5090 GPU\footnote{Our code will be publicly available after acceptance.}.
We use SMART~\cite{yan2025advancing} to measure execution performance in a realistic simulation environment. SMART is a physics-based, ADG-driven testbed that models continuous robot kinodynamics, low-level controller behavior, network communication, and inter-agent interaction effects, while scaling to thousands of agents.
These properties make it well suited for benchmarking MAPF planners under realistic execution conditions.
Following the motion model commonly adopted in warehouse robotics and MAPF deployments, each robot translates only along its current direction and changes direction through in-place rotation.
As shown in~\Cref{fig:exp:sim_env}, we evaluate on eight maps converted from the MovingAI Benchmark~\cite{Stern2019benchmark}: \texttt{empty} (empty-32-32, $32 \times 32$), \texttt{maze} (maze-32-32-4, $32 \times 32$), \texttt{random} (random-64-64-10, $64 \times 64$), \texttt{room} (room-64-64-16, $64 \times 64$), \texttt{den312d} ($65 \times 81$), \texttt{WH-S} (warehouse-10-20-10-2-1, $161 \times 63$), \texttt{WH-L} (warehouse-20-40-10-2-1, $321 \times 123$), and \texttt{boston} (Boston\_0\_256, $256 \times 256$), where each grid cell corresponds to a $1 \times 1$\,m workspace region.
For each map, we use 25 instances.
Each instance specifies start-goal pairs and per-agent kinodynamic
parameters: a maximum linear velocity in $[1, 4]$\,m/s, linear acceleration in $[0.5, 1.0]$\,m/s$^2$, and maximum angular velocity in $[\pi/4, \pi]$\,rad/s, sampled from a correlated distribution.

The remainder of this section is organized as follows.
\Cref{sec:exp:predictor} evaluates the prediction accuracy of \mypredictor on
in- and out-of-distribution maps and agent counts.
\Cref{sec:exp:planning} evaluates \mymethod integrated with MAPF-LNS and CBS.
\Cref{sec:exp:esadg} evaluates \myexecution against execution-ordering
baselines, followed by a case study on MAPF with deadlines
(\Cref{sec:exp:case_study}). Finally, \Cref{sec:exp:real_robot} reports
real-robot validation results.

\subsection{Evaluation on \mypredictor} \label{sec:exp:predictor}
We first describe the training procedure for \mypredictor, then evaluate it on in- and out-of-distribution data, and finally conduct an ablation study to assess feature importance.

% describe the training steps
\subsubsection{Data Collection and Model Training}
To train \mypredictor, we first collect MAPF solutions from six maps: \texttt{empty}, \texttt{maze}, \texttt{random}, \texttt{room}, \texttt{den312d}, and \texttt{WH-S}. For each map, we use MAPF-LNS~\cite{li2021anytime} to plan all 25 random instances with agent counts ranging from 10 to 300, saving the plan at every iteration to obtain multiple MAPF solutions from each run. These plans are executed in SMART to obtain execution information as training labels. Simultaneously, we extract encoded graphs from the ADGs constructed for each plan to use as input features. In total, we collect 28,177 graphs for training.

\mypredictor follows a hybrid GNN–LSTM architecture that alternates between temporal and spatial processing. Node features are first grouped into per-agent sequences and processed by a 3-layer unidirectional LSTM with hidden dimension 32 to capture sequential dependencies along each agent’s path. The resulting node representations are then passed through three stacked GATv2 graph attention layers (hidden dimension 32, ReLU activations) to aggregate spatial context across inter-agent dependency edges. A second 3-layer LSTM (hidden dimension 32) re-integrates the spatially enriched representations in temporal order, after which a 3-layer MLP predicts the per-node arrival time.
The model is trained by minimizing the mean absolute percentage error (MAPE) between predicted and ground-truth arrival times.
To provide additional supervisory signal, an auxiliary MLP branch that shares the same backbone predicts the arrival speed at each node.
We use the AdamW optimizer with an initial learning rate of $10^{-3}$, weight decay of $10^{-4}$, and a step learning rate schedule that halves the rate every 10 epochs. Training runs for up to 300 epochs with a batch size of 8 and early stopping if validation MAPE does not improve for 20 consecutive epochs.

% Compare our performance both on training maps (with new paths) and new maps (generalizability) w/ different path length (diffent map size), different num of agents
% baseline 1: GNN-VAE, baseline 2: ConvLSTM
\subsubsection{Results}

\begin{table}[t]
\small
\caption{MAPE (\%) across (1) GNNExecNet, (2) LSTMExecNet, and (3) \mypredictor. We mark out-of-distribution maps with (*)}
\centering
{
\begin{tabular}{clccc}
\toprule
 {Map} & {$M$} & (1) & (2) & (3) \\
\midrule
\multirow{7}{*}{\rotatebox{90}{\texttt{empty}}} 
& 50  & $8.2 \pm 0.7$  & $5.4 \pm 0.6$  & \textbf{2.75 $\pm$ 0.20} \\
& 100 & $10.1 \pm 0.7$ & $6.7 \pm 0.5$  & \textbf{3.37 $\pm$ 0.28} \\
& 150 & $10.7 \pm 0.5$ & $6.9 \pm 0.3$  & \textbf{3.64 $\pm$ 0.21} \\
& 200 & $10.9 \pm 0.6$ & $7.0 \pm 0.4$  & \textbf{3.83 $\pm$ 0.18} \\
& 300 & $10.5 \pm 0.4$ & $6.7 \pm 0.3$  & \textbf{3.96 $\pm$ 0.16} \\
& 400 & $9.9 \pm 0.4$  & $6.6 \pm 0.3$  & \textbf{4.12 $\pm$ 0.21} \\
& 500 & $8.9 \pm 0.4$  & $6.2 \pm 0.2$  & \textbf{3.90 $\pm$ 0.14} \\
\midrule

\multirow{7}{*}{\rotatebox{90}{\texttt{WH-S}}} 
& 50  & $8.4 \pm 0.8$  & $6.5 \pm 0.7$  & \textbf{2.21 $\pm$ 0.28} \\
& 100 & $10.1 \pm 0.6$ & $7.9 \pm 0.5$  & \textbf{2.89 $\pm$ 0.21} \\
& 150 & $10.5 \pm 0.5$ & $8.2 \pm 0.4$  & \textbf{3.31 $\pm$ 0.25} \\
& 200 & $11.0 \pm 0.5$ & $8.4 \pm 0.4$  & \textbf{3.68 $\pm$ 0.33} \\
& 300 & $11.6 \pm 0.5$ & $9.0 \pm 0.4$  & \textbf{4.20 $\pm$ 0.22} \\
& 400 & $12.5 \pm 0.7$ & $9.1 \pm 0.5$  & \textbf{4.56 $\pm$ 0.28} \\
& 500 & $13.3 \pm 0.6$ & $8.6 \pm 0.3$  & \textbf{4.63 $\pm$ 0.20} \\
\midrule

\multirow{6}{*}{\rotatebox{90}{\texttt{random}}} 
& 50  & $8.6 \pm 0.8$  & $5.8 \pm 0.7$  & \textbf{2.86 $\pm$ 0.19} \\
& 100 & $9.9 \pm 0.7$  & $6.8 \pm 0.4$  & \textbf{3.47 $\pm$ 0.34} \\
& 150 & $10.1 \pm 0.5$ & $6.9 \pm 0.4$  & \textbf{3.70 $\pm$ 0.18} \\
& 200 & $10.0 \pm 0.4$ & $6.7 \pm 0.3$  & \textbf{3.84 $\pm$ 0.19} \\
& 300 & $9.5 \pm 0.4$  & $6.5 \pm 0.3$  & \textbf{4.02 $\pm$ 0.16} \\
& 400 & $8.7 \pm 0.4$  & $6.2 \pm 0.3$  & \textbf{3.95 $\pm$ 0.21} \\
\midrule

\multirow{7}{*}{\rotatebox{90}{\texttt{WH-L}*}} 
& 50  & $8.6 \pm 1.3$  & $7.1 \pm 1.1$  & \textbf{3.20 $\pm$ 0.32} \\
& 100 & $9.1 \pm 0.7$  & $8.8 \pm 0.7$  & \textbf{3.41 $\pm$ 0.29} \\
& 150 & $9.7 \pm 0.5$  & $9.8 \pm 0.6$  & \textbf{3.49 $\pm$ 0.22} \\
& 200 & $10.4 \pm 0.4$ & $10.5 \pm 0.6$ & \textbf{3.48 $\pm$ 0.18} \\
& 300 & $11.2 \pm 0.4$ & $11.3 \pm 0.3$ & \textbf{3.65 $\pm$ 0.16} \\
& 400 & $11.7 \pm 0.3$ & $11.7 \pm 0.4$ & \textbf{3.76 $\pm$ 0.16} \\
& 500 & $12.3 \pm 0.4$ & $12.1 \pm 0.4$ & \textbf{3.93 $\pm$ 0.15} \\
\midrule

\multirow{7}{*}{\rotatebox{90}{\texttt{boston}*}} 
& 50  & $9.7 \pm 0.9$  & $8.2 \pm 0.8$  & \textbf{2.98 $\pm$ 0.43} \\
& 100 & $9.7 \pm 0.5$  & $9.3 \pm 0.8$  & \textbf{3.39 $\pm$ 0.29} \\
& 150 & $9.5 \pm 0.3$  & $10.4 \pm 0.5$ & \textbf{4.01 $\pm$ 0.29} \\
& 200 & $9.7 \pm 0.3$  & $10.9 \pm 0.6$ & \textbf{4.44 $\pm$ 0.41} \\
& 300 & $10.1 \pm 0.4$ & $12.0 \pm 0.6$ & \textbf{5.14 $\pm$ 0.45} \\
& 400 & $10.4 \pm 0.3$ & $12.6 \pm 0.4$ & \textbf{5.49 $\pm$ 0.35} \\
& 500 & $10.7 \pm 0.3$ & $12.9 \pm 0.4$ & \textbf{5.91 $\pm$ 0.37} \\

\bottomrule
\end{tabular}
}
\label{table:mape}
\end{table}

In this section, we evaluate the performance of \mypredictor by comparing it with two baseline methods: (1) GNNExecNet, inspired by~\cite{meng2025reliable}, uses a graph neural network with two GATv2 layers to predict execution time based on the ADG, and (2) LSTMExecNet, follows~\cite{ge2023congestion} and processes each agent’s action sequence independently using LSTM models to predict execution times.
Performance is measured using MAPE, which reports the prediction error as a percentage of the actual execution time.
A single trained model is used across all evaluation settings without any map-specific or scale-specific fine-tuning.
We evaluate prediction accuracy under two settings: (1) \textit{In-distribution}: we use new paths from the training maps to test interpolation, and (2)~\textit{Out-of-distribution}, where we test on two unseen maps (\texttt{WH-L} and \texttt{boston}) that differ in both topology and scale.
Within each setting, we vary the number of agents from 50 to 500, exceeding the maximum training count of 300, so that results at 400 and 500 agents further probe out-of-distribution generalization along the agent-count axis.
As shown in~\Cref{table:mape}, \mypredictor consistently achieves significantly lower MAPE than all baselines.
Moreover, its performance remains stable when transitioning from in-distribution to out-of-distribution scenarios or when varying the number of agents, demonstrating strong generalization across both topology and scale.
Beyond accuracy, \mypredictor is fast enough for use within search: a single SMART rollout on \texttt{WH-L} with 300 agents takes about 27 s, whereas \mypredictor returns a prediction in 0.3 s (\cref{table:runtime}), nearly two orders of magnitude faster.

\subsubsection{Ablation Study}

\begin{table}[t]
\centering
\small
\caption{Feature importance analysis: impact of removing action, graph structure, and edge features.}
\resizebox{\linewidth}{!}{\begin{tabular}{lcc}
\toprule
{Feature Configuration} & {MAPE (\%)} & {$\Delta$ MAPE (\%)} ↓  \\
\midrule
Full Model (All Features)       & 3.80 & -  \\
w/o Action Features             & 3.91 & 0.11  \\
w/o Graph Structural Features    & 4.94 &  1.14 \\
w/o Edge Features               & 4.57 &  0.77 \\
w/o Per-agent Kinodynamic Features & 5.35 & 1.55 \\
\bottomrule
\end{tabular}}
\label{tab:ablation}
\end{table}

To evaluate the contribution of different feature groups, we conduct an ablation study by training \mypredictor with each group removed in turn.
As shown in~\Cref{tab:ablation}, removing per-agent kinodynamic features causes the largest degradation, confirming that heterogeneous motion capabilities are the single most informative signal for predicting execution times.
Graph structural features rank second, reflecting their role in encoding how delays propagate through the dependency graph.
Edge features contribute a further $+0.77\%$, while action features have a smaller but still positive effect. All four groups improve accuracy when present, indicating that our selected features capture complementary sources of execution-time variation.

\subsection{Evaluation on \mymethod} \label{sec:exp:planning}

\begin{figure*}
    \centering
    \includegraphics[width=\linewidth]{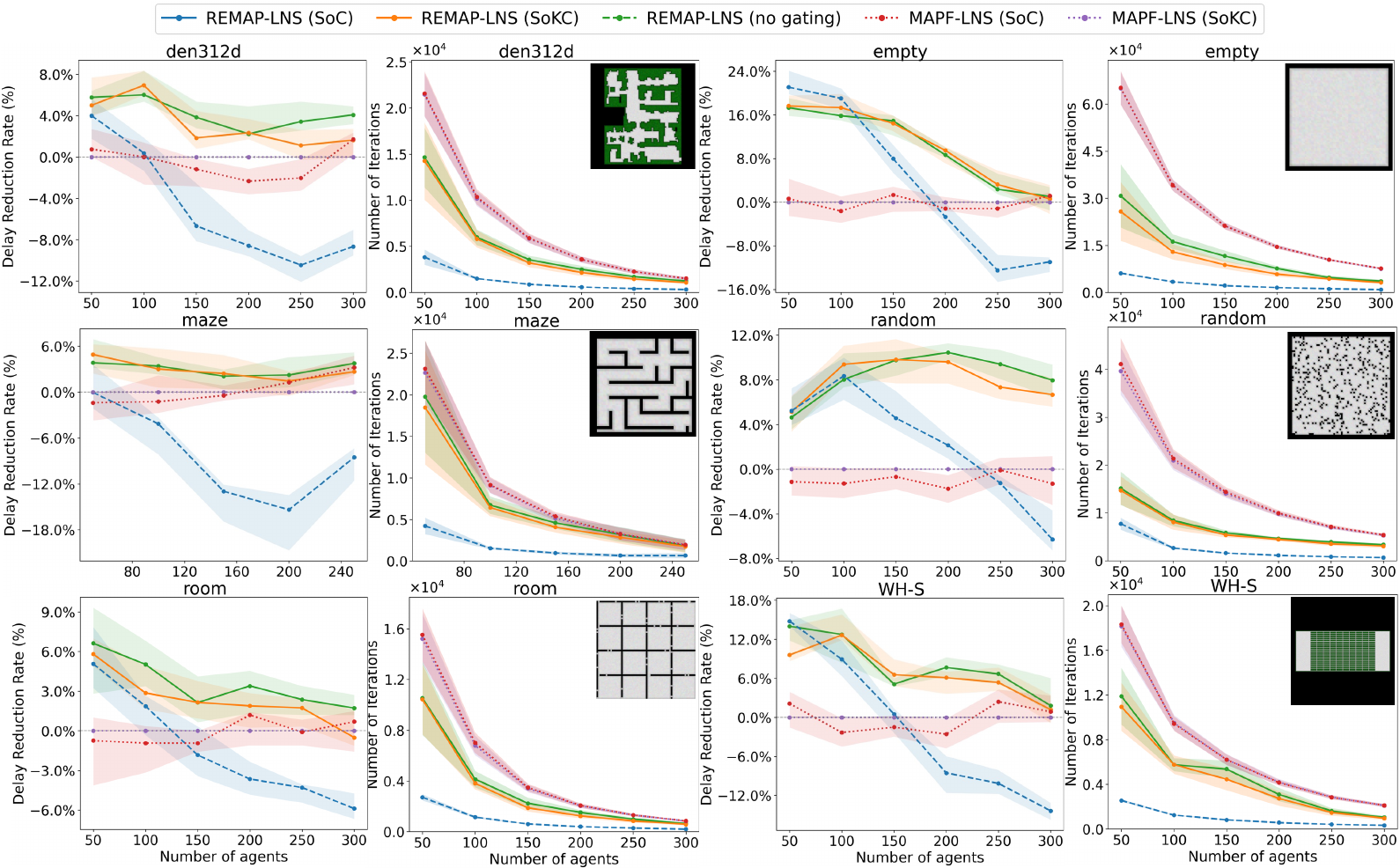}
    \caption{Results of REMAP-LNS compared with MAPF-LNS. DRR (left, higher is better) and number of LNS iterations completed within the time budget (right) across six maps. Shaded regions indicate standard deviation across 25 scenarios.}
    \label{fig:exp2_lns_main}
\end{figure*}

We evaluate \mymethod on six MovingAI Benchmark maps (\texttt{empty}, \texttt{maze}, \texttt{random}, \texttt{room}, \texttt{den312d}, and \texttt{WH-S}) with 25 random instances per map.
We first integrate \mymethod with MAPF-LNS and then with CBS.

\subsubsection{Experiment Setup}
We first compare methods built on MAPF-LNS~\cite{li2021anytime}:
(1)~\emph{MAPF-LNS (SoC)}: the standard MAPF-LNS baseline that accepts or rejects candidates based solely on the sum of costs;
(2)~\emph{MAPF-LNS (SoKC)}: a variant that replaces the sum of costs with the kinodynamic-aware cost $J_{\text{SoKC}} = \sum_i T_i / \bar{u}_i$, where $T_i$ is the path length and $\bar{u}_i$ is the maximum speed of agent $a_i$, providing a stronger proxy for execution time but still without querying \mypredictor;
(3)~\emph{REMAP-LNS (no gating)}: queries \mypredictor at every LNS iteration to evaluate $J_{\text{exec}}$, representing the most accurate but most expensive way to evaluate execution cost;
(4)~\emph{REMAP-LNS (SoC)}: uses the adaptive gating mechanism (\Cref{sec:remap:lns}) with the standard sum of costs as the cheap objective $J_{\text{fast}}$; and
(5)~\emph{REMAP-LNS (SoKC)}: uses adaptive gating with $J_{\text{SoKC}}$ as the cheap objective, combining the stronger proxy with selective oracle queries.

We report solution quality using the \emph{Delay Reduction Rate} (DRR):
\begin{equation} \label{eq:gcr}
\text{DRR} = \frac{J_{\text{exec}}^{\text{base}} - J_{\text{exec}}^{\text{method}}}{J_{\text{exec}}^{\text{base}} - J_{\text{exec}}^{\text{ideal}}},
\end{equation}
where $J_{\text{exec}}^{\text{base}}$ is the execution cost of the \emph{MAPF-LNS (SoC)} baseline, $J_{\text{exec}}^{\text{method}}$ is the cost of the evaluated method, and $J_{\text{exec}}^{\text{ideal}} = \sum_i d_i / \bar{u}_i$ is an idealized reference summing each agent's shortest-path traversal time at maximum speed $\bar{u}_i$.
A higher DRR means the method removes more of the avoidable delay.
Since $J_{\text{exec}}^{\text{ideal}}$ assumes conflict-free, congestion-free travel, it falls well below any cost attainable in practice, so the denominator overstates the removable delay. The reported DRR values are therefore conservative: the true reductions relative to the attainable optimum exceed these percentages.

\subsubsection{Results}
\Cref{fig:exp2_lns_main} reports DRR (odd columns) and the number of LNS iterations completed within the time budget (even columns) across all six maps.
The cheap-objective baselines fail to improve on the standard planner.
\emph{MAPF-LNS (SoC)} sits at the $0\%$ reference by construction, and \emph{\mymethod-LNS (SoKC)} remains within a narrow band around it. Despite completing the largest number of iterations on every map, neither baseline yields meaningful improvement, indicating that the mismatch between discrete planning cost and true execution cost cannot be closed by substituting one cheap proxy for another.
Querying \mypredictor at every iteration is harmful at scale. 
\emph{\mymethod-LNS (no gating)} attains the strongest per-iteration quality at low agent counts, reaching up to $20\%$ DRR at 50 agents on \texttt{empty}, but DRR drops sharply with agent count and becomes substantially negative on five of the six maps, falling below $-10\%$ at 300 agents on \texttt{den312d}, \texttt{maze}, \texttt{empty}, and \texttt{WH-S}.
The iteration plots show an order of magnitude fewer iterations than the cheap-objective baselines, confirming that the planner is starved of iterations and cannot reach competitive solutions despite a more faithful per-candidate objective.
\emph{\mymethod-LNS (SoKC)} and \emph{\mymethod-LNS (SoC)} maintain positive DRR across nearly all map and agent-count combinations, with the two variants tracking each other closely
throughout. On \texttt{random}, both sustain DRR above $7\%$ from 50 to 300 agents, while \emph{\mymethod-LNS (no gating)} crosses into negative territory beyond 200 agents on the same map.
The iteration plots show that the adaptive variants complete roughly $2\times$ the iterations of \emph{\mymethod-LNS (no gating)} while remaining below the cheap-objective baselines.
The gating mechanism thus realizes the intended tradeoff, sustaining high solution quality across agent counts and remaining competitive with MAPF-LNS under the fast objective as agent count grows and the runtime budget tightens.

\begin{table}[t]
\caption{Runtime breakdown. Columns 3 to 6 correspond to the number of nodes in the ADG,  time for ADG construction, \mypredictor inference time, and total time, respectively.}
\centering
\resizebox{\linewidth}{!}{
\begin{tabular}{cccccc}
\toprule
 {Map} & \multicolumn{1}{c}{{$M$}} & Nodes \# & \multicolumn{1}{c}{ADG build (ms)} & Inference (ms) & Total (ms) \\ 
\midrule
\multirow{3}{*}{\rotatebox{90}{empty}}
& 50 & 2241 & 0.4 $\pm$ 0.0 & 6.6 $\pm$ 0.1 & 7.1 $\pm$ 0.1 \\ %\cmidrule(lr){2-6}
& 100 & 5000 & 1.0 $\pm$ 0.0 & 12.5 $\pm$ 0.1 & 13.5 $\pm$ 0.1 \\ %\cmidrule(lr){2-6}
& 300 & 16279 & 6.1 $\pm$ 0.2 & 49.4 $\pm$ 0.5 & 55.5 $\pm$ 0.7 \\ %\cmidrule(lr){2-6}
\midrule
% \multirow{3}{*}{\rotatebox{90}{WH-S}}
% & 50 & 8665 & 1.5 $\pm$ 0.0 & 14.6 $\pm$ 0.1 & 16.1 $\pm$ 0.1 \\ %\cmidrule(lr){2-6}
% & 100 & 18986 & 3.9 $\pm$ 0.2 & 33.8 $\pm$ 0.3 & 37.6 $\pm$ 0.3 \\ %\cmidrule(lr){2-6}
% & 300 & 52125 & 16.9 $\pm$ 0.2 & 133.8 $\pm$ 0.8 & 150.7 $\pm$ 1.0 \\ %\cmidrule(lr){2-6}
% \midrule
\multirow{3}{*}{\rotatebox{90}{WH-L}}
& 50 & 18003 & 3.8 $\pm$ 0.1 & 32.6 $\pm$ 0.1 & 36.4 $\pm$ 0.3 \\ %\cmidrule(lr){2-6}
& 100 & 38270 & 9.9 $\pm$ 0.4 & 69.5 $\pm$ 0.9 & 79.4 $\pm$ 1.0 \\ %\cmidrule(lr){2-6}
& 300 & 116345 & 38.3 $\pm$ 0.3 & 277.5 $\pm$ 2.3 & 315.8 $\pm$ 2.3 \\ %\cmidrule(lr){2-6}
\bottomrule
\end{tabular}
}
\label{table:runtime}
\end{table}

\subsubsection{Runtime Analysis}
We report the time required for \mymethod to construct the ADG and perform inference in~\cref{table:runtime}. \mymethod efficiently constructs the ADG and performs inference with reasonable computational overhead. 
While these processing times increase with agent count, they remain sufficiently fast to support the planner effectively. Under ideal conditions, even on the WH-L map with 300 agents, our method can support approximately 200 iterations in MAPF-LNS within 1 minute.
These results suggest promising opportunities for further optimization.
As shown in~\Cref{fig:exp2_lns_main}, \mymethod consistently outperforms the cheap-objective baselines when iteration counts are matched.
Reducing the cost of \mypredictor inference therefore represents a direct pathway to further improving \mymethod at large agent populations.

\begin{figure}
    \centering
    \includegraphics[width=1.\linewidth]{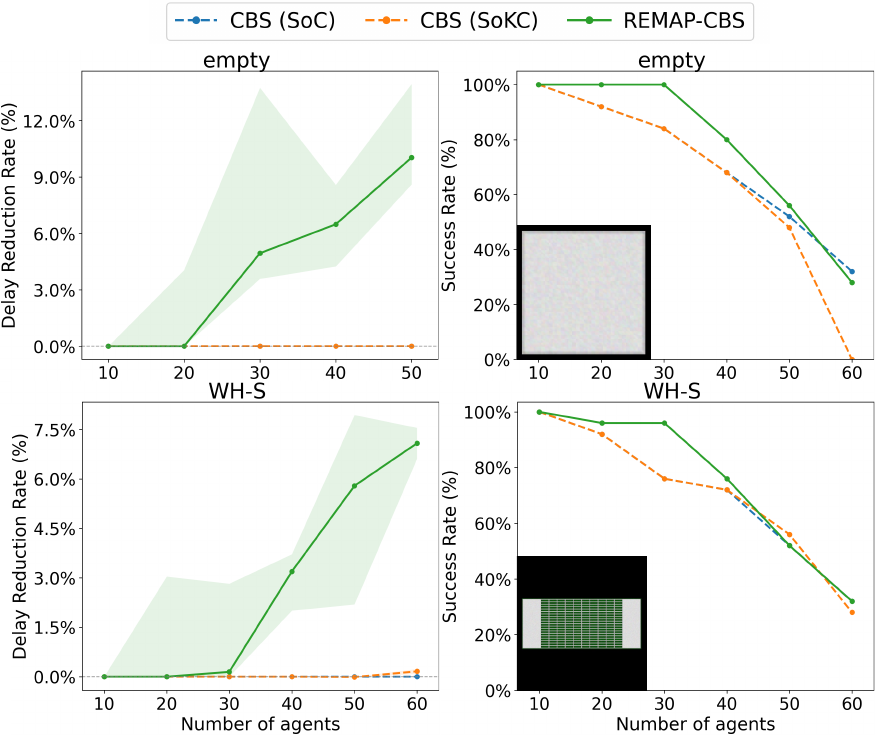}
    \caption{Results of \mymethod integrated with CBS. The left column reports the DRR achieved by \mymethod relative to the baseline CBS planner, while the right column reports planner success rate.}
    \label{fig:exp2_cbs_results}
\end{figure}

\subsubsection{Integration with CBS}
To show the performance of \mymethod with planners belonging to the CBS family, we additionally integrate the framework with CBS~\cite{sharon2015conflict}, a representative optimal MAPF planner.
The experiments are conducted on the \texttt{empty} and \texttt{WH-S} maps with progressively increasing numbers of agents. For each setting, we evaluate 25 instances with a one-minute planning time limit.
We compare three methods: (1)~\emph{CBS (SoC)}, the standard CBS planner with the sum of costs as the objective; 
(2)~\emph{CBS (SoKC)}, which replaces the sum of
costs with the kinodynamic-aware cost; and (3)~\emph{\mymethod-CBS}, which uses the estimations from \mypredictor as the objective.
We report planner success rate, defined as the percentage of instances solved within the time limit, and the DRR defined in~\Cref{eq:gcr}.

Results are shown in \Cref{fig:exp2_cbs_results}. On both maps, \mymethod consistently improves execution performance over the baseline CBS planner, and the improvement becomes more pronounced as the number of agents increases. On the \texttt{empty} map, \mymethod achieves 10.3\% improvement in terms of DRR at 50 agents.
Similar trends are observed on \texttt{WH-S}. 
At the same time, the planner success rate remains close to the baseline across most settings. This is partly because CBS is already computationally expensive, so the additional overhead introduced by execution-time reasoning has relatively limited impact on the overall planning time.

\subsection{Evaluation on \myexecution}\label{sec:exp:esadg}
In this section, we first compare the performance of \myexecution with two state-of-the-art baseline methods.
Then, we run an ablation study to isolate the contribution of the edge selection heuristic.

\subsubsection{Experiment Setting}
We evaluate \myexecution on the same six MovingAI Benchmark maps and kinodynamic setup described in~\Cref{sec:exp:planning}, with 25 scenarios per map and varying agent counts.
Initial plans are generated using 1-robust PBS~\cite{ma2019searching,atzmon2020robust}, which ensures cycle-conflict-free solutions suitable for ADG construction.
For \myexecution, we use a top-$K$ candidate size of 10, and a wall-clock cutoff of 60\,s.
All optimized schedules are executed in SMART to confirm that predicted improvements translate to actual gains.
We compare against two optimal methods that also optimize execution orderings on fixed paths:
SADG~\cite{berndt2023receding}, which solves the SADG reordering problem via mixed-integer linear programming, and
STPG~\cite{JiangAAAI25}, a dedicated search algorithm with improved scalability.
We quantify optimization quality using the \emph{Normalized Improvement Ratio} (NIR):
\begin{equation}
  \text{NIR} = \frac{T_{\text{init}} - T_{\text{post}}}{T_{\text{init}} - T_{\text{lb}}},
  \label{eq:nir}
\end{equation}
where $T_{\text{init}}$ is the execution time of the initial unoptimized schedule, $T_{\text{post}}$ is the post-optimization execution time, and $T_{\text{lb}}$ is a lower bound computed as the sum of each agent's individual optimal execution time under the assumption of no inter-agent waiting.
This bound underestimates the true minimum, since the assumption is generally infeasible in the presence of shared locations.

\begin{figure}
    \centering
    \includegraphics[width=\linewidth]{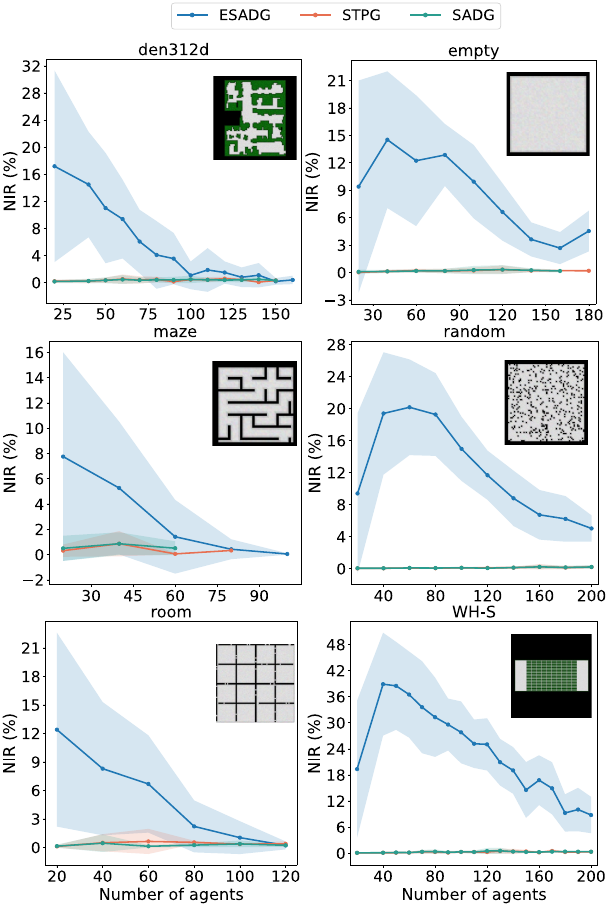}
    \caption{Normalized Improvement Ratio (NIR) of \myexecution, STPG, and SADG across six maps with varying agent counts. Higher NIR indicates a larger fraction of the gap between the unoptimized schedule and the theoretical lower bound has been closed. Shaded regions indicate standard deviation.}
    \label{fig:exp3_ksadg_main}
\end{figure}

\begin{figure}[t]
    \centering
    \includegraphics[width=.9\linewidth]{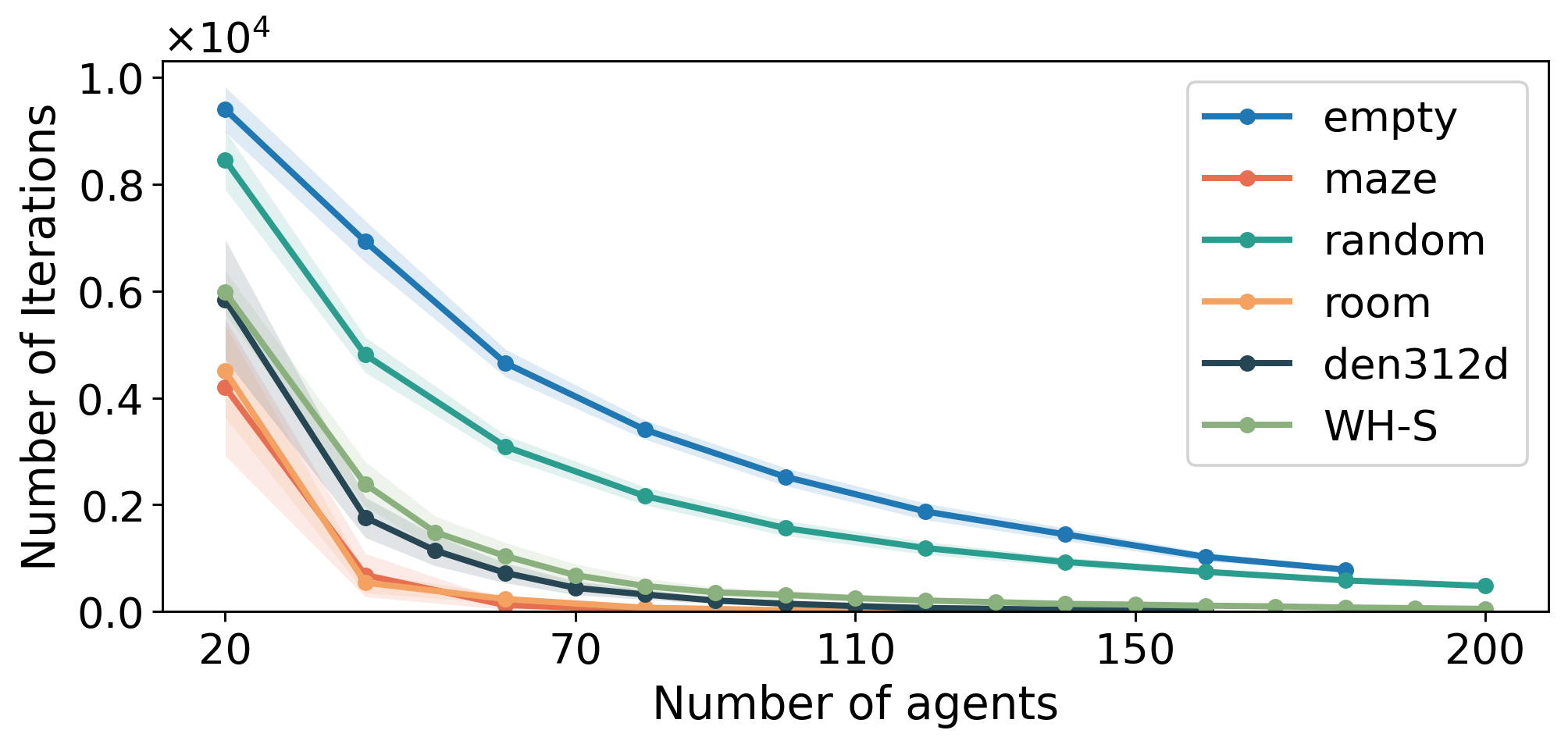}
    \caption{Number of \myexecution iterations completed within the 60\,s time budget as a function of agent count. Larger ADGs require more expensive predictor calls, sharply reducing the number of orderings explored.}
    \label{fig:ksadg_iterations}
\end{figure}

\subsubsection{Results}

\Cref{fig:exp3_ksadg_main} reports NIR across all six maps.
\myexecution consistently and substantially outperforms both baselines across all maps and agent counts, achieving NIR values of 10--40\% where STPG and SADG remain near zero.
The strongest improvements appear on \texttt{WH-S}, where \myexecution reaches up to 40\% NIR, reflecting the large number of inter-agent conflicts in warehouse corridors that create significant room for reordering.
Maps with more complex topology (\texttt{random}, \texttt{den312d}) also show large gains, peaking around 20\% and 17\% respectively.
NIR tends to decrease as agent count grows, as shown in~\Cref{fig:ksadg_iterations}, this is primarily because a larger MAPF plan require more expensive predictor calls and feasibility checks per iteration, sharply reducing the number of iterations \myexecution can complete within the fixed time budget.
At 20 agents, the search explores up to 10{,}000 iterations; at 200 agents, this drops to fewer than 1{,}000 on most maps.
With fewer iterations, the \myexecution search covers a smaller fraction of the feasible ordering space, limiting the quality of the solution found.
This suggests that reducing the cost per-iteration through faster inference is a direct pathway to improving \myexecution's performance at scale.
The near-zero NIR of STPG and SADG highlights the advantage of using a learned surrogate.
These methods rely on a simplified objective, making it hard to optimize to a high quality solution.

\begin{figure}
    \centering
    \includegraphics[width=\linewidth]{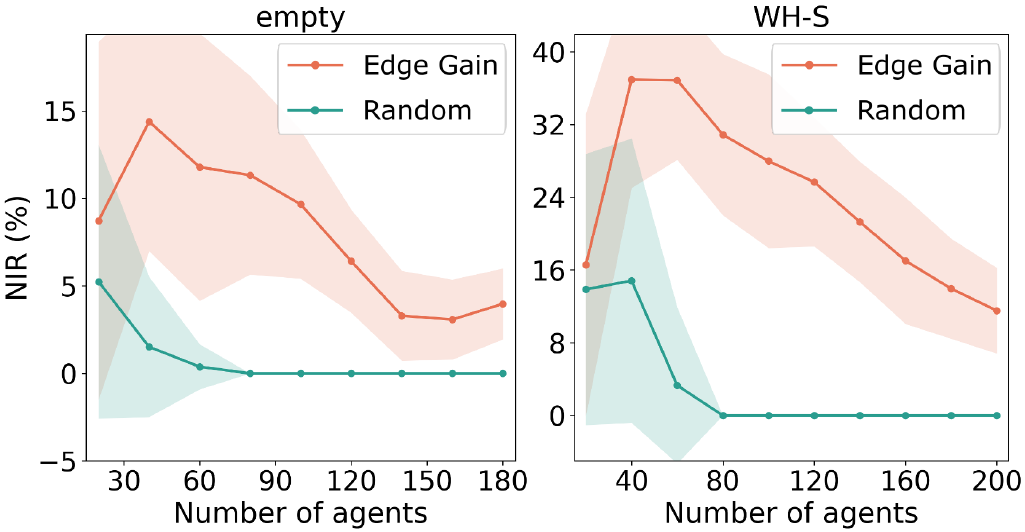}
    \caption{Ablation study on reverse edge selection heuristic.}
    \label{fig:exp2_ablation}
\end{figure}

\subsubsection{Ablation Study}
To isolate the contribution of the edge selection strategy described in~\cref{sec:ksadg:selection}, we compare two variants: \emph{Edge Gain}, which uses the gain-based ranking from~\cref{sec:ksadg:selection} as the default \myexecution configuration, and \emph{Random}, which replaces gain-based ranking with uniform random selection among non-tabu groups while keeping all other components identical.
\Cref{fig:exp2_ablation} reports NIR for both variants on \texttt{empty} and \texttt{WH-S}.

On both maps, the gain estimator provides a substantial advantage.
On \texttt{WH-S}, \emph{Edge Gain} peaks at approximately $38\%$ NIR around 40 agents, while \emph{Random} reaches only $14\%$ at the same scale. The gap is equally pronounced on \texttt{empty}, where \emph{Edge Gain} sustains $14\%$ NIR at low agent counts compared to $5\%$ for \emph{Random}. More notably, \emph{Random} degrades to zero NIR beyond 80 agents on both maps, whereas \emph{Edge Gain} retains meaningful improvements even at 200 agents on both maps. 
As agent count grows, the fixed time budget permits fewer iterations, making each iteration's choice of reversal group increasingly consequential. Without the gain estimator, the search expends its limited budget on low-impact reversals and fails to locate beneficial reorderings.

\subsection{Case Study: MAPF with Real-World Deadlines}\label{sec:exp:case_study}

We now evaluate both components of our framework, \mymethod and \myexecution, on MAPF with Real-world Deadlines (MAPF-RD)~\cite{ma2018mapfdl,huang2023deadline}, a representative time-sensitive application where each agent must reach its goal before a specified deadline.

\subsubsection{Problem Formulation}\label{exp:Prbo}
MAPF-RD arises as a specific instantiation of execution-aware MAPF.
Suppose each agent $a_i$ is assigned a goal location and a deadline $T_i^D \in \mathbb{R}_{+}$.
Each agent needs to reach its goal location before its deadline.
We define the following \emph{deadline miss ratio} function:

\[
J_{\text{deadline}}(\pi)
=
\frac{1}{M}
\sum_{i=1}^{M}
\mathbb{I}[\tau_i(\pi) > T_i^D].
\]

Here, $\mathbb{I}[\cdot]$ denotes the indicator function, which returns 1 when agent $a_i$ misses its deadline (i.e., $\tau_i(\pi) > T_i^D$) and 0 otherwise.
It counts missed deadlines and focuses on the overall task completion rate.
The objective of MAPF-RD is to compute a collision-free solution $\mathcal{P}$ that minimizes the deadline miss ratio which is a special case of $J_{\text{exec}}$.

\begin{figure}
    \centering
    \includegraphics[width=\linewidth]{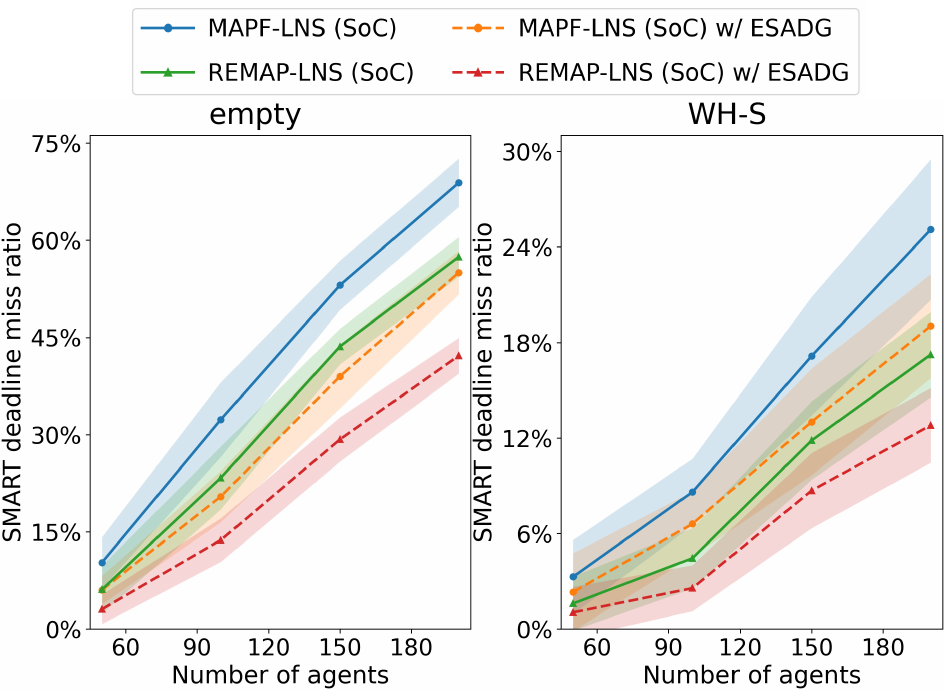}
    \caption{Deadline miss ratio on \texttt{empty} and \texttt{WH-S} as a
    function of agent count. Lower is better. Shaded regions indicate
    standard deviation.}
    \label{fig:exp3_mapfdl}
\end{figure}

\subsubsection{Experiment Setup}
We conduct this case study on two maps, \texttt{empty} and \texttt{WH-S} with 25 scenarios per map and agent counts varying from 50 to 200. Per-agent deadlines are obtained by multiplying each agent's lower-bound execution time, defined as the shortest path length divided by the maximum speed, by a scaling factor sampled uniformly from $[K_D, 3 \cdot K_D]$. 
All plans are executed in SMART to obtain wall-clock arrival times for penalty evaluation.
We compare four methods that vary along two axes: whether planning uses the discrete sum-of-costs objective (\emph{MAPF-LNS}) or our execution-aware objective (\emph{\mymethod-LNS}), and whether \myexecution is applied as a subsequent execution-time optimization. This yields \emph{MAPF-LNS (SoC)}, \emph{\mymethod-LNS (SoC)}, and their counterparts with \myexecution, \emph{MAPF-LNS (SoC) w/ ESADG} and \emph{\mymethod-LNS (SoC) w/ ESADG}. The design isolates the contribution of each layer and reveals how planning quality interacts with execution-ordering optimization.

\subsubsection{Results}
\Cref{fig:exp3_mapfdl} reports the deadline miss ratio across both maps. Every method shows a rising miss ratio with agent count as congestion increases, though the magnitude differs substantially.
The discrete-objective baseline \emph{MAPF-LNS (SoC)} performs worst on both maps, reaching $69\%$ on \texttt{empty} and $25\%$ on \texttt{WH-S} at 200 agents. This reflects a mismatch between the planning objective and the execution results: sum-of-costs ignores the wall-clock delays incurred during execution, so plans that look improved under the planning objective do not always reduce deadline violations.
Replacing the planning objective with execution-time predictions addresses this directly. \emph{\mymethod-LNS (SoC)} lowers the miss ratio at every agent count, by up to $12\%$ on \texttt{empty} and $8\%$ on \texttt{WH-S} relative to \emph{MAPF-LNS (SoC)}, confirming that execution-aware planning is central to MAPF-RD.
Applying \myexecution yields further gains on top of either planner. On its own it is effective: \emph{MAPF-LNS (SoC) w/ ESADG} improves markedly over \emph{MAPF-LNS (SoC)} and approaches the execution-aware planner. Combining both layers performs best: \emph{\mymethod-LNS (SoC) w/ ESADG} attains the lowest miss ratio at every agent count on both maps, reaching $42\%$ on \texttt{empty} and $13\%$ on \texttt{WH-S} at 200 agents, a reduction of $27\%$ and $12\%$ respectively over \emph{MAPF-LNS (SoC)}.
The two layers contribute through distinct mechanisms: execution-aware planning shapes the paths, while \myexecution arbitrates shared locations during execution, and their combination compounds the individual gains.

\subsection{Evaluation on Real Robots} \label{sec:exp:real_robot}

\begin{figure}
    \centering
    \includegraphics[width=1.\linewidth]{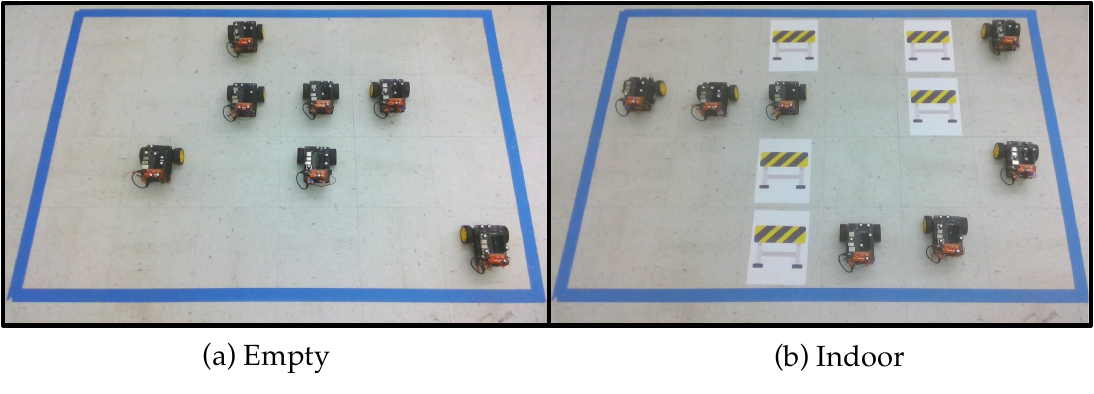}
    \caption{Real-robot experimental environments.}
    \label{fig:real_world_env}
\end{figure}

To validate that the predictions of \mypredictor trained in simulation transfer to physical systems and that \mymethod and \myexecution improve execution-time performance on real hardware, we conduct experiments on a fleet of real robots.

\subsubsection{Experiment Setup}
We deploy 7 NVIDIA Jetbot robots on a $4 \times 6$ grid workspace, where each cell measures $0.3 \times 0.3$\,m.
As shown in~\cref{fig:real_world_env}, we consider two map configurations to introduce varying degrees of topological complexity and inter-agent congestion: an \emph{empty} layout with no obstacle and an \emph{indoor} layout with narrow passages that increase agent interactions.
A Vicon motion capture system provides centralized localization at 100\,Hz, and a central server runs the MAPF planner, constructs the ADG, and dispatches actions to the robots.
For each configuration, we generate 10 problem instances by sampling random start-goal pairs for all agents.
Each instance is evaluated under three methods: (1)~MAPF-LNS alone; (2)~MAPF-LNS with \mymethod; and (3)~MAPF-LNS with \mymethod followed by \myexecution.

The physical platform differs from SMART in controller dynamics, grid scale, communication latency, and localization characteristics, which induces a domain gap for the simulation-trained \mypredictor (\Cref{sec:exp:predictor}).
To address this, we collect a calibration dataset of 100 executions distributed evenly across both map configurations by running MAPF-LNS plans on the physical system and recording actual wall-clock arrival times. The resulting graphs are partitioned into 80 for training and 20 for testing. Starting from the simulation-pretrained weights, \mypredictor is fine-tuned for 100 epochs at a learning rate of $10^{-3}$.

\subsubsection{Results}

\begin{table}[t]
\small
\caption{Sum of execution time across (1)~MAPF-LNS;
(2)~MAPF-LNS with \mymethod; and
(3)~MAPF-LNS with \mymethod followed by \myexecution post-processing in seconds.}
\centering
% \resizebox{\linewidth}{!}
{
\begin{tabular}{lccc}
\toprule
 {Map} & (1) & (2) & (3) \\
\midrule
\texttt{empty} 
& $111.9 \pm 15.4$  & $98.0 \pm 22.6$  & $94.8 \pm 21.2$ \\
\midrule
\texttt{indoor}
& $207.4 \pm 72.3$  & $198.7 \pm 66.3$  & $193.2 \pm 69.2$ \\
\bottomrule
\end{tabular}
}
\label{table:real_robots}
\end{table}

\Cref{table:real_robots} reports the sum of execution time across all agents for each method on both map configurations.
On the \texttt{empty} map, MAPF-LNS with \mymethod reduces the total execution time from $111.9$\,s to $98.0$\,s, a relative improvement of 12.4\% over the baseline. Applying \myexecution on top further lowers the total to $94.8$\,s, yielding a combined reduction of 15.3\% relative to MAPF-LNS alone. On the \texttt{indoor} map, the improvements are more modest in relative terms: \mymethod reduces execution time by 4.2\%, and the full pipeline achieves a 6.8\% reduction overall. The smaller gains on \texttt{indoor} can be attributed to the narrow passages, which constrain the space of feasible reorderings and limit the planner's ability to reroute agents through alternative paths.
Across both configurations, \myexecution provides a consistent additional reduction beyond what \mymethod achieves during planning, confirming that planning optimization and execution ordering optimization address complementary sources of inefficiency on physical hardware, consistent with the findings from simulation experiments.

Regarding sim-to-real transfer, the fine-tuned \mypredictor achieves a MAPE of $9.3\%$ on the held-out test set of 20 physical executions. While this error is higher than the simulation-only results reported in~\Cref{table:mape}, the gap is expected for two reasons.
First, the Jetbot platform exhibits substantially higher control noise and hardware variability, which introduces stochastic execution-time variation that the predictor cannot fully resolve from the MAPF plan alone.
In industrial deployments, however, this source of error is expected to be considerably smaller.
Second, the calibration dataset is limited to 100 executions due to laboratory constraints, whereas large-scale operators with access to fleet-level logging can readily collect orders of magnitude more data for fine-tuning.
Despite this prediction error, \mymethod and \myexecution still produce measurable execution-time reductions on the physical system, suggesting that the framework is tolerant of moderate predictor inaccuracy and that prediction accuracy is likely to improve further in deployment settings where richer calibration data is available.

\section{Conclusion} \label{sec:conclusion}
This paper presented an execution-aware framework for MAPF that bridges the gap between discrete planning objectives and real-world execution performance. We introduced ExecTimeNet, a dependency-graph-based world model to predict realized execution times and states from MAPF solutions. Building on this model, we proposed REMAP, which incorporates execution-time estimates into MAPF search, and an adaptive gating mechanism that reduces prediction overhead by selectively invoking the model. We further introduced ESADG, an execution-time optimization method that refines inter-agent precedence orderings to improve execution performance while preserving path feasibility and deadlock-free execution.
Experiments in realistic simulation and on physical robots show that the proposed framework improves execution performance over methods based only on discrete planning objectives. We also study deadline satisfaction as a representative special case of execution-aware MAPF, demonstrating that reasoning about predicted wall-clock arrival times can improve performance under practical timing constraints. These results suggest that learned execution-time models can provide an effective bridge between discrete MAPF planning and realistic multi-robot execution. Future work will extend the framework to additional execution-aware objectives and broader robot platforms.

% Future work: 1. deveolop learning method that better integrate the planning method so it can find solution in a sense of real world execution

\bibliographystyle{IEEEtran}
\bibliography{reference}

% \include{appendix}

% \newpage

 

% \end{IEEEbiography}



% \vfill

\end{document}